\documentclass{article}
\pdfoutput=1

     \PassOptionsToPackage{numbers, compress}{natbib}

 \usepackage[dblblindworkshop,final]{neurips_2025}
\usepackage[utf8]{inputenc} %
\usepackage[T1]{fontenc}    %
\usepackage{hyperref}       %
\usepackage{url}            %
\usepackage{booktabs}       %
\usepackage{amsfonts}       %
\usepackage{nicefrac}       %
\usepackage{microtype}      %
\usepackage{xcolor}         %
\usepackage{overpic}
\usepackage{ulem}

\usepackage{mauraisMacros2}
\usepackage{mathtools}
\usepackage[capitalize,noabbrev]{cleveref}

\title{Learning Paths for Dynamic Measure Transport: A Control Perspective} 
\workshoptitle{Frontiers in Probabilistic Inference: Sampling Meets Learning}

\author{%
  Aimee Maurais \\
  Massachusetts Institute of Technology\\
  Cambridge, MA 02139 \\
  \texttt{maurais@mit.edu} \\
  \And
  Bamdad Hosseini \\
  University of Washington \\
  Seattle, WA 98195 \\
  \texttt{bamdadh@uw.edu} \\
  \AND
  Youssef Marzouk \\
  Massachusetts Institute of Technology\\
  Cambridge, MA 02139 \\
  \texttt{ymarz@mit.edu} 
}

\usepackage[textsize=tiny]{todonotes}
\definecolor{darkgreen}{rgb}{.15,.55,0}

\newcommand{\rev}[1]{\textcolor{black}{#1}}

\graphicspath{{figures/}}

\begin{document}

\maketitle

\begin{abstract}
We bring a control perspective to the problem of identifying {paths of measures} for sampling via dynamic measure transport (DMT). We highlight the fact that commonly used paths may be poor choices for DMT and connect existing methods for learning alternate paths to mean-field games. Based on these connections we pose a flexible family of optimization problems for identifying {tilted} paths of measures for DMT and advocate for the use of objective terms which encourage {smoothness} of the corresponding velocities. We present a numerical algorithm for solving these problems based on recent Gaussian process methods for solution of partial differential equations 
and demonstrate the ability of our method to recover more efficient and smooth transport models compared to those which use an untilted reference path.
\end{abstract}

\section{Introduction}
{Sampling} from a target probability distribution $\pi \in \calP(\R^d)$ is a fundamental task in modern machine learning, enabling, e.g., %
uncertainty quantification in Bayesian inference \citep{ghanem2017handbook} and generation of convincing synthetic data  \citep{cao2024survey,papamakarios2021normalizing,creswell2018generative}. %
Many recent sampling algorithms %
are grounded in a 
\textit{dynamic measure transport (DMT)} framework, which typically makes use of %
a stochastic differential equation (SDE) 
\begin{equation}
\rmd X_t = v(X_t, t) \, \rmd t + \sigma \rmd W_t, \quad t \in [0, T], \quad X_0 \sim \eta,
\label{eq:dynamics}
\end{equation}
where $v: \R^d \times [0, T] \to \R^d$ is the \textit{drift} or \textit{velocity}, $\sigma \geq 0$ is a fixed noise level, $W_t$ is white noise, $\eta$ is a reference measure,
and $T > 0$ is a stopping time. %
Broadly speaking, \rev{the} goal is to design the dynamics \eqref{eq:dynamics} such that $X_T \sim \pi$. In practice, due 
to limitations of data and computation, we ask for 
an approximate process $\widehat{X}_t$ such that
${\rm Law} (\widehat{X}_T) \approx \pi$. 
This can often be cast as a learning problem for an approximate 
drift $\widehat{v} \approx v$.
With $\widehat{v}$ in hand, 
we can generate approximate samples from $\pi$
by simulating \eqref{eq:dynamics} with $\widehat{v}$ 
to transform samples from $\eta$ into approximate 
samples from $\pi$.

The SDE \eqref{eq:dynamics} (an ODE for $\sigma \!=\! 0$) induces a \textit{path of distributions} $(\rho(t))_{t \in [0, T]}$, where 
$\rho(t) = \mathrm{Law}(X_t)$, satisfying 
$\rho(0) = \eta$ and $\rho(T) = \pi$. %
In some DMT approaches, such as neural ODEs and continuous normalizing flows \citep{chenNeuralOrdinaryDifferential2018,grathwohlFFJORDFreeFormContinuous2018}, this path is implicit or of little concern, but in 
more recent methods, such as diffusion models or stochastic 
interpolants \citep{songScoreBasedGenerativeModeling2021,albergoStochasticInterpolantsUnifying2023,mauraisSamplingUnitTime2024c,chemseddineNeuralSamplingBoltzmann2024,wangMeasureTransportKernel2024}, the path 
is explicit and at the heart of the methodology. 
In these latter methods the drift $\widehat{v}$ is identified not only such that $\mathrm{Law}(\widehat{X}_T) \approx \pi$, \textit{but also such that $\mathrm{Law}(\widehat{X}_t) \approx \rho(t)$, for all
$t \in [0,T]$}. 
As $\rho$ and $v$ 
must jointly satisfy a Fokker--Planck equation (FPE) corresponding to \eqref{eq:dynamics}, the entire problem of DMT can be cast as one of approximately solving the FPE; some recent techniques are based precisely on this 
idea \cite{sunDynamicalMeasureTransport2024a,mauraisSamplingUnitTime2024c,mateLearningInterpolationsBoltzmann2023a}.

\section{Good and bad paths of measures}
In this article we consider the following question: 
\begin{quote}
\it
    Can we identify a problem-dependent path of densities $\rho(t)$
for which 
    an associated drift $v$ and sample trajectories $X_t$ can be well approximated?  
\end{quote} %
Our motivation stems from the fact that some DMT approaches can be used with virtually \textit{any} tractable path of measures so long as the required ``ingredients'' for approximating $v$ are available. For instance, stochastic interpolants \citep{albergoStochasticInterpolantsUnifying2023,albergoBuildingNormalizingFlows2022} use paths given by the law of a random variable interpolation which can be constructed rather arbitrarily. Likewise, density-driven DMT approaches often use the geometric annealing path between $\eta$ and $\pi$, but there are some, e.g., \citep{mateLearningInterpolationsBoltzmann2023a,mauraisSamplingUnitTime2024c,sunDynamicalMeasureTransport2024a,wangMeasureTransportKernel2024}, that could, in principle, be used with any path of measures with an accessible log-derivative. %
Within these flexible frameworks it is not often clear which paths are best, especially given that canonical paths like the McCann interpolant \citep{mccannConvexityPrincipleInteracting1997} are typically intractable.
The current practice in DMT approaches that allow a choice of path is seemingly to choose \rev{one} which is easy to write down: in stochastic interpolants \citep{albergoBuildingNormalizingFlows2022,lipmanFlowMatchingGenerative2022,liuFlowStraightFast2022a} the default path corresponds to a linear interpolation between reference and target random variables, and in density-driven settings practitioners tend to %
employ the geometric annealing path. 

\subsection{Issues with the geometric annealing path} 
The geometric annealing path, given by $\mu(t) \propto \eta^{1-t}\pi^t$, $t \in [0,1]$ is convenient for density-driven DMT because %
it has a log-derivative which is independent of normalizing constants. Moreover, it possesses Fisher--Rao gradient flow structure \citep{domingo-enrichExplicitExpansionKullbackLeibler2023,chenGradientFlowsSampling2023} and variational characterizations (e.g., \citep[Theorem 4.9]{amari2016information}). %
It may, however, be problematic for DMT with certain combinations of $\eta$ and $\pi$. %
This issue was, to our knowledge, first highlighted in \citet{mateLearningInterpolationsBoltzmann2023a}. We demonstrate this phenomenon via the example $\eta = \calN(0, 1)$ and $\pi = \frac{2}{3}\calN(-8, 1) + \frac{1}{3}\calN(4, 1)$ in the top row of \cref{fig:learned_interp_1D}. %
\newcommand{\sfwidth}{0.0825\linewidth}
\begin{figure}[h]
\begin{subfigure}{\sfwidth}
	\centering 
    \subcaption*{\footnotesize $t=0$}
    \begin{overpic}[width=\linewidth]{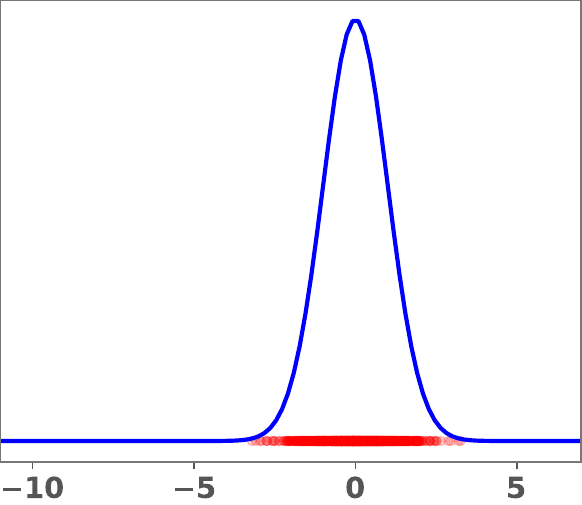}
        \put(-20,18){\rotatebox{90}{\tiny Geometric}}
    \end{overpic}
\end{subfigure} 
\begin{subfigure}{\sfwidth}
	\centering 
    \subcaption*{\footnotesize $t=0.1$}
	\includegraphics[width=\linewidth]{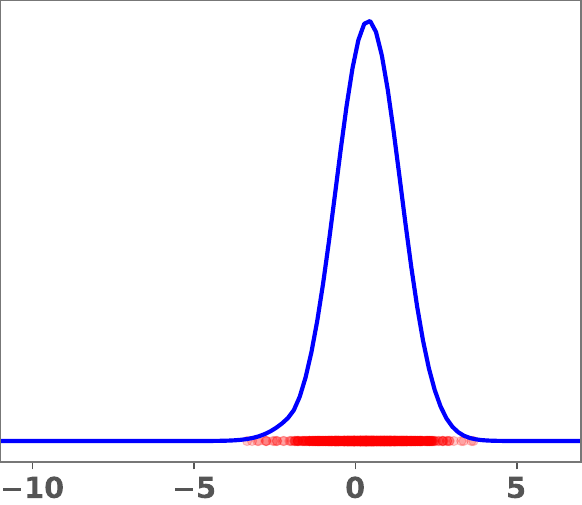}
\end{subfigure} 
\begin{subfigure}{\sfwidth}
	\centering 
    \subcaption*{\footnotesize $t=0.2$}
	\includegraphics[width=\linewidth]{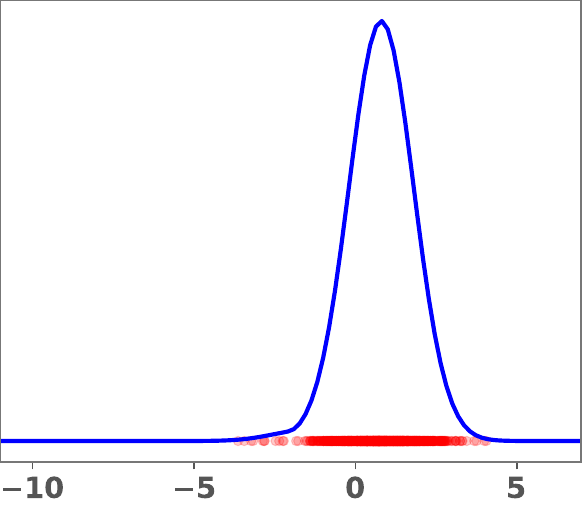}
\end{subfigure} 
\begin{subfigure}{\sfwidth}
	\centering 
    \subcaption*{\footnotesize $t=0.3$}
	\includegraphics[width=\linewidth]{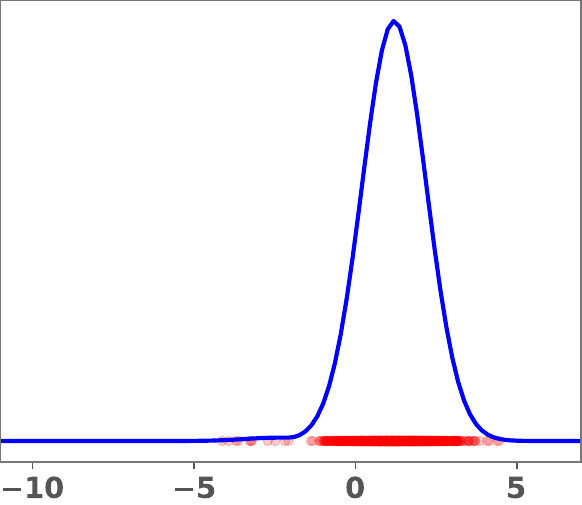}
\end{subfigure} 
\begin{subfigure}{\sfwidth}
	\centering 
    \subcaption*{\footnotesize $t=0.4$}
	\includegraphics[width=\linewidth]{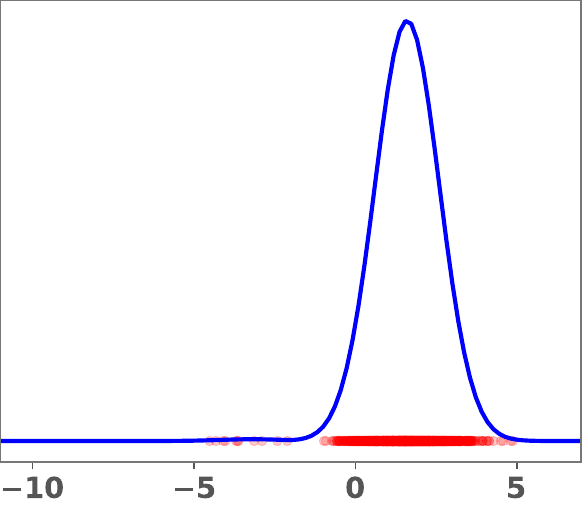}
\end{subfigure} 
\begin{subfigure}{\sfwidth}
	\centering 
    \subcaption*{\footnotesize $t=0.5$}
	\includegraphics[width=\linewidth]{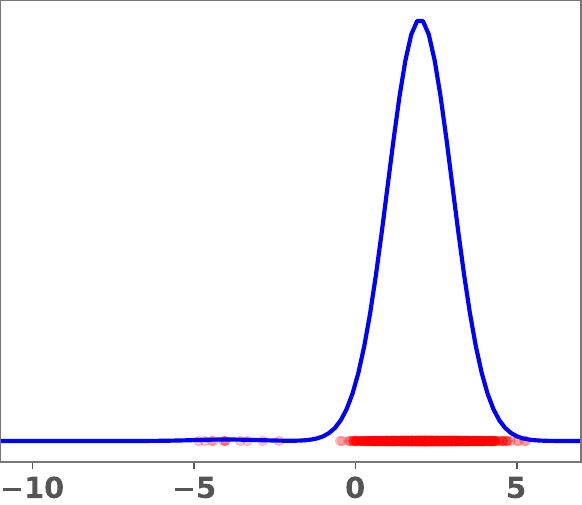}
\end{subfigure} 
\begin{subfigure}{\sfwidth}
	\centering 
    \subcaption*{\footnotesize $t=0.6$}
	\includegraphics[width=\linewidth]{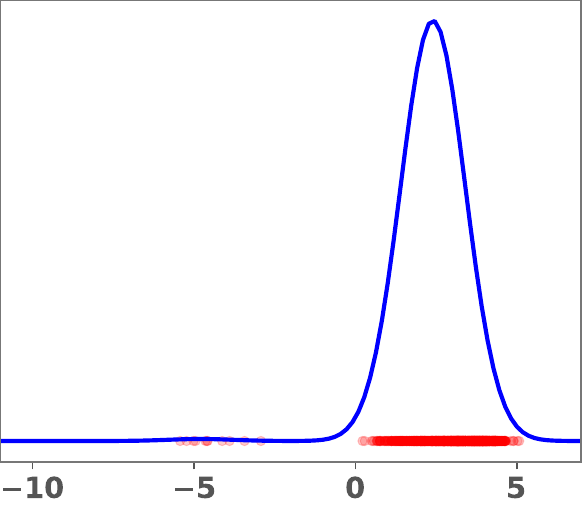}
\end{subfigure} 
\begin{subfigure}{\sfwidth}
	\centering 
    \subcaption*{\footnotesize $t=0.7$}
	\includegraphics[width=\linewidth]{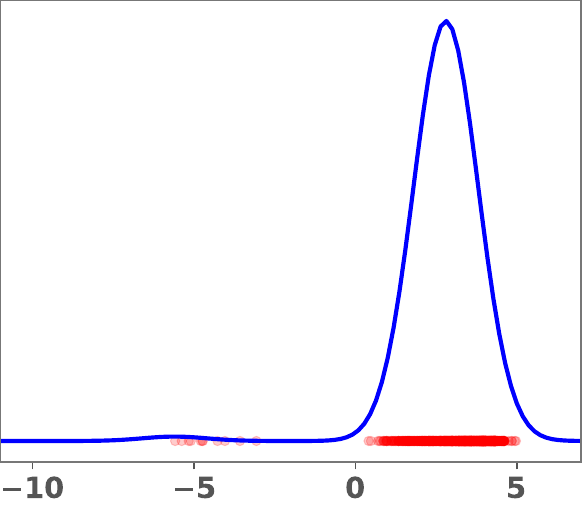}
\end{subfigure} 
\begin{subfigure}{\sfwidth}
	\centering 
    \subcaption*{\footnotesize $t=0.8$}
	\includegraphics[width=\linewidth]{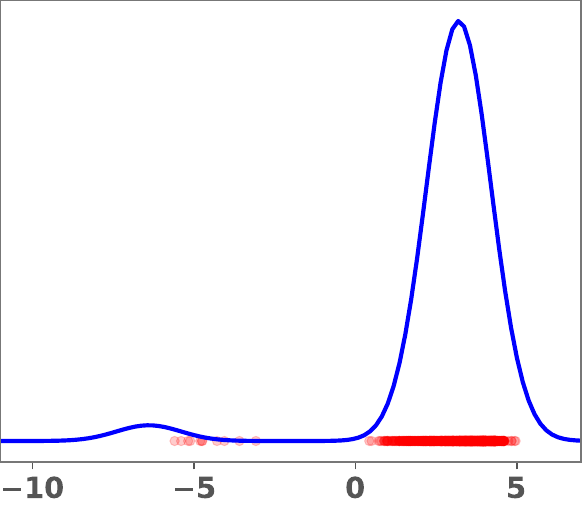}
\end{subfigure} 
\begin{subfigure}{\sfwidth}
	\centering 
    \subcaption*{\footnotesize $t=0.9$}
	\includegraphics[width=\linewidth]{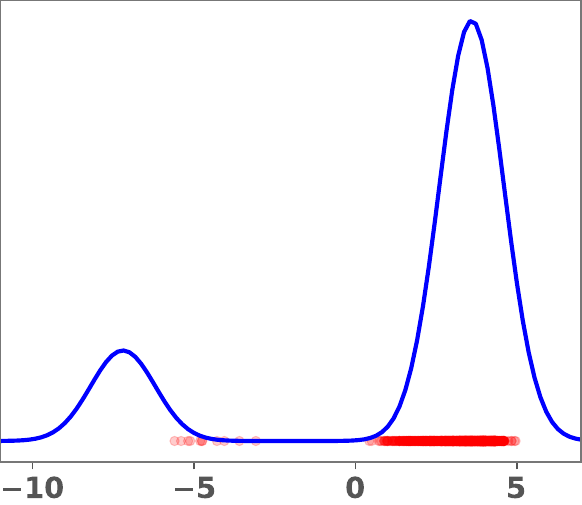}
\end{subfigure} 
\begin{subfigure}{\sfwidth}
	\centering 
    \subcaption*{\footnotesize $t=1.0$}
	\includegraphics[width=\linewidth]{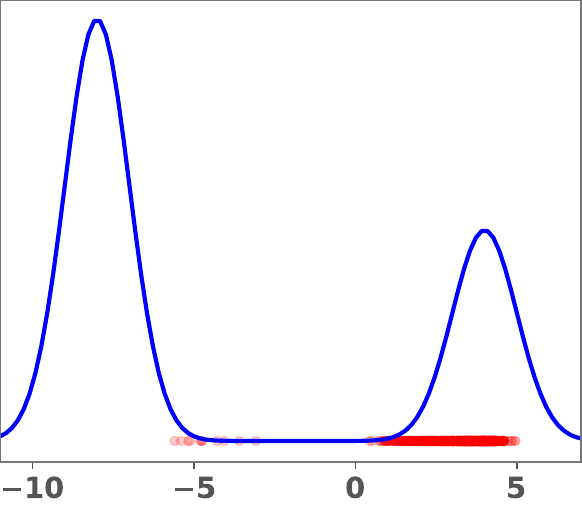}
\end{subfigure} 
\\ 
\begin{subfigure}{\sfwidth}
	\centering 
    \begin{overpic}[width=\linewidth]{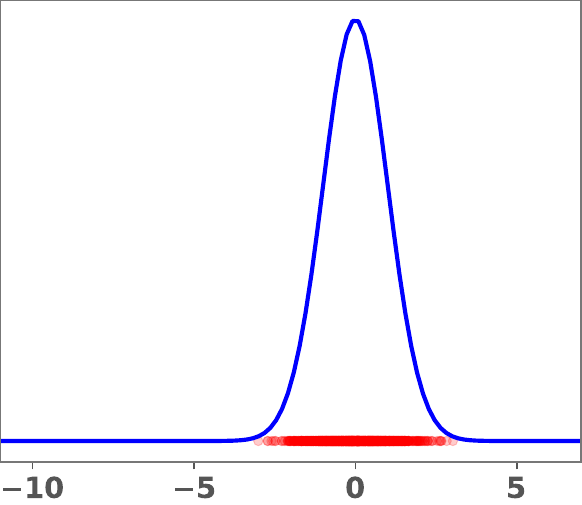}
        \put(-20,18){\rotatebox{90}{\tiny Learned}}
    \end{overpic}
\end{subfigure} 
\begin{subfigure}{\sfwidth}
	\centering 
	\includegraphics[width=\linewidth]{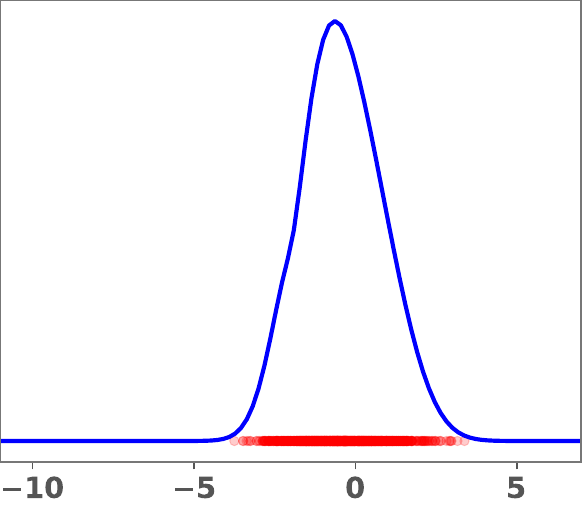}
\end{subfigure} 
\begin{subfigure}{\sfwidth}
	\centering 
	\includegraphics[width=\linewidth]{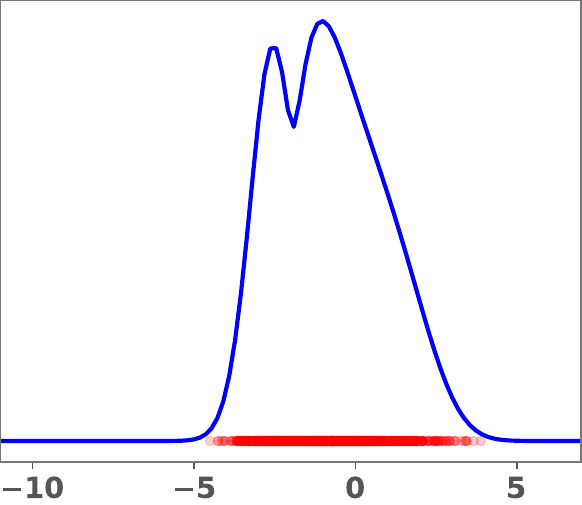}
\end{subfigure} 
\begin{subfigure}{\sfwidth}
	\centering 
	\includegraphics[width=\linewidth]{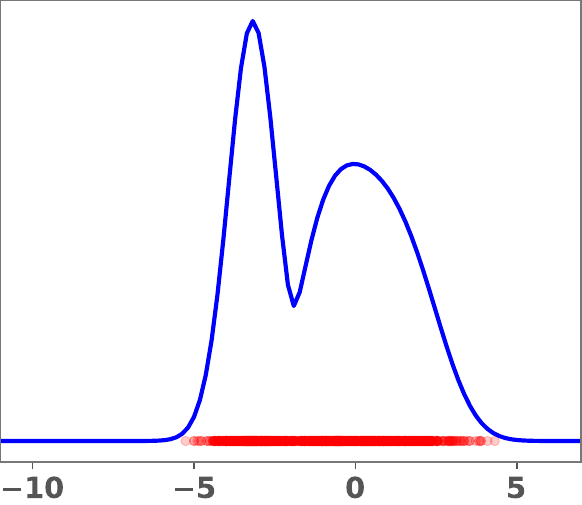}
\end{subfigure} 
\begin{subfigure}{\sfwidth}
	\centering 
	\includegraphics[width=\linewidth]{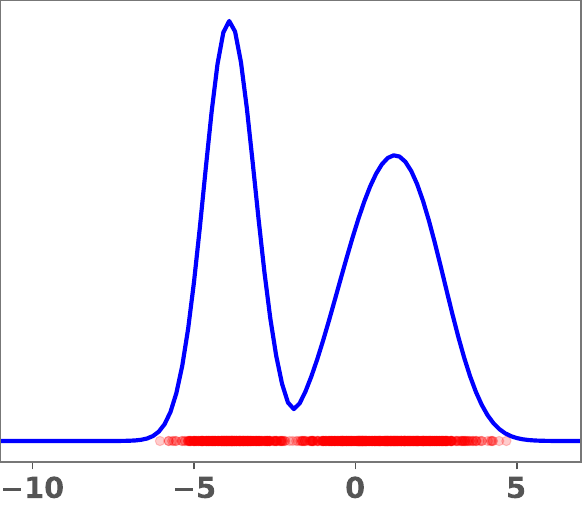}
\end{subfigure} 
\begin{subfigure}{\sfwidth}
	\centering 
	\includegraphics[width=\linewidth]{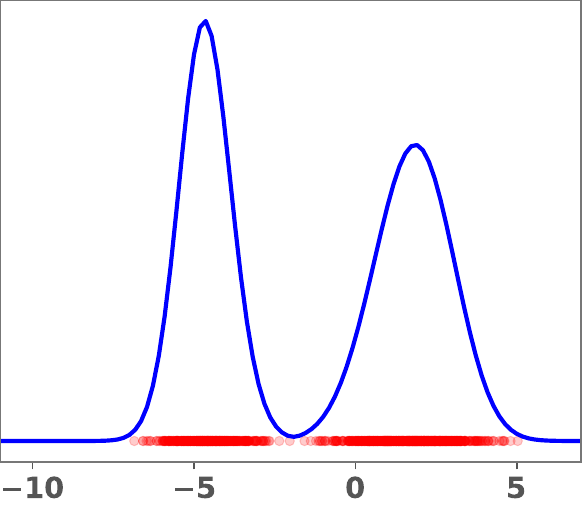}
\end{subfigure} 
\begin{subfigure}{\sfwidth}
	\centering 
	\includegraphics[width=\linewidth]{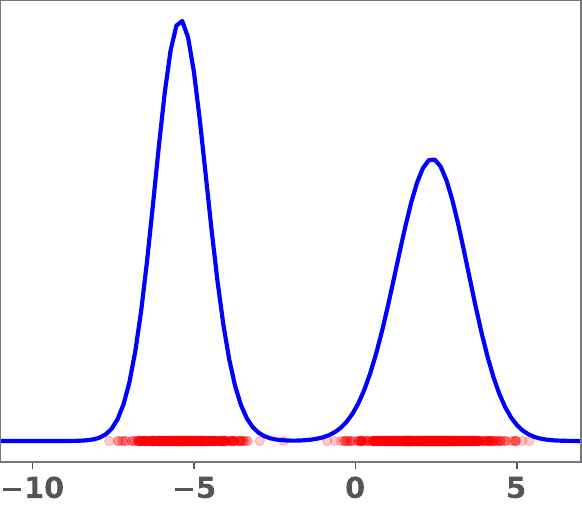}
\end{subfigure} 
\begin{subfigure}{\sfwidth}
	\centering 
	\includegraphics[width=\linewidth]{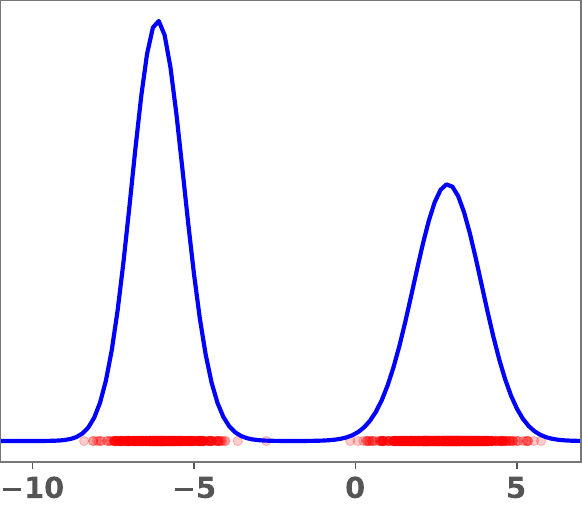}
\end{subfigure} 
\begin{subfigure}{\sfwidth}
	\centering 
	\includegraphics[width=\linewidth]{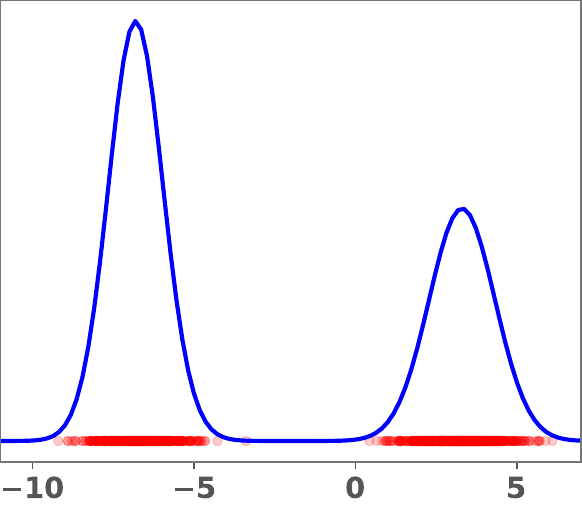}
\end{subfigure} 
\begin{subfigure}{\sfwidth}
	\centering 
	\includegraphics[width=\linewidth]{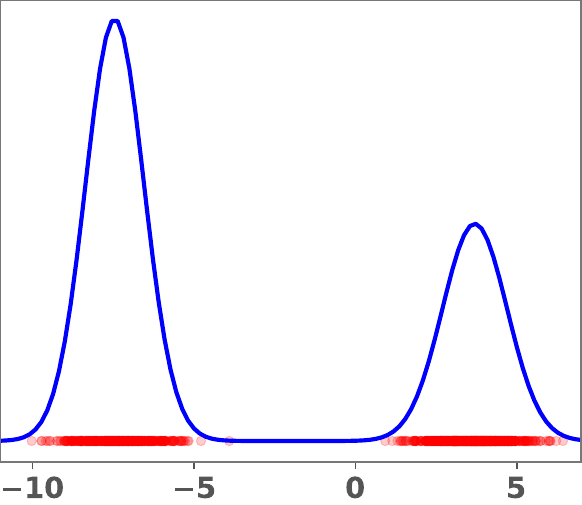}
\end{subfigure} 
\begin{subfigure}{\sfwidth}
	\centering 
	\includegraphics[width=\linewidth]{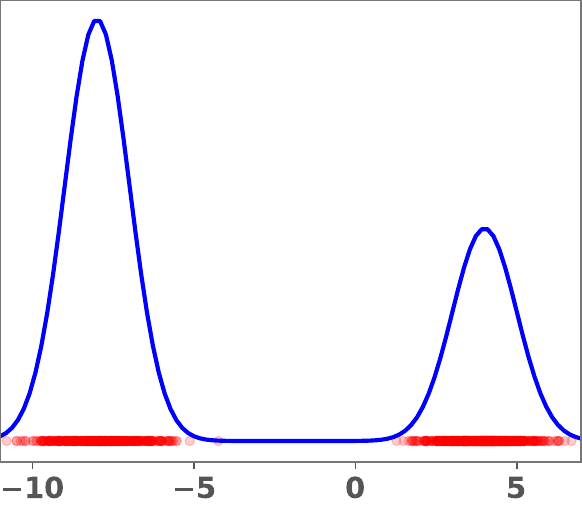}
\end{subfigure} 
    \caption{Geometric annealing path (top) and path resulting from solving  our proposed control problem \eqref{eq:relaxed_problem} (bottom) for the example $\eta = \calN(0, 1)$ and $\pi = \frac{2}{3}\calN(-8, 1) + \frac{1}{3}\calN(4, 1)$. Samples generated by the respective velocity fields are plotted overtop in red.}
\label{fig:learned_interp_1D}
\end{figure}
The evolution of $\mu(t)$ is dominated by transport from $\eta$ to the closest mode of $\pi$ until $t \approx 0.8$, at which point ``teleportation of mass'' from the lesser to the greater mode begins. Capturing this teleportation with DMT is difficult; the velocity we identify by numerically solving the FPE (see \cref{sec:numerics}) almost completely fails to place samples in the left mode. The physics-informed neural network (PINN) approach used in \citet{mateLearningInterpolationsBoltzmann2023a} faced similar difficulties with analogous examples. 
Even if an algorithm could learn a velocity achieving transport along $\mu(t)$ for this $(\eta, \pi)$, such a velocity would be large and irregular; see  \citet{chemseddineNeuralSamplingBoltzmann2024} for results in this vein. 

\subsection{A fix and an explanation}
\label{sec:fix_and_mfg}
The approach taken in \citet{mateLearningInterpolationsBoltzmann2023a} to correct  teleportation behavior of $\mu(t)$ is to add a \textit{perturbation} $f: \R^d \times [0, 1] \to \R$ to the log of the geometric mixture, 
\begin{equation}
\log \mu^f(\cdot, t)  = (1 - t) \log \eta(\cdot) + t \log \pi(\cdot) + t(1 - t) f(\cdot, t) - \log Z(t),
\label{eq:log_mut_f}
\end{equation}
where $Z(t) = \int_{\R^d} \eta^{1 - t} \pi^t e^{t(1 - t) f(\cdot, t)} \, \rmd x$ is the normalizing constant. The interpolation \eqref{eq:log_mut_f} ensures $\mu^f(0) = \eta$ and $\mu^f(1) = \pi$ and corresponds to a {tilting} $\mu^f(\cdot, t) \propto \mu(\cdot, t)e^{t(1 - t)f(\cdot, t)}$ of $\mu$. In \cite{mateLearningInterpolationsBoltzmann2023a}, $f$ is learned alongside a velocity field $v$ by minimizing a PINN loss corresponding to the continuity equation for ODE transport along the  path \eqref{eq:log_mut_f}. 
This optimization problem is strongly ill-posed---there are infinitely many $f$s one could use in \eqref{eq:log_mut_f}, and even for fixed $f$ there are infinitely many valid velocities $v$. Yet, remarkably, the $f$ and $v$ \citep{mateLearningInterpolationsBoltzmann2023a} recovers are quite well-behaved \citep[Figure 8]{mateLearningInterpolationsBoltzmann2023a}. %

Obtaining a nice path by minimizing a PINN loss over neural networks is not a given; in replicating the results of \citep{mateLearningInterpolationsBoltzmann2023a} we found that considerable tuning was necessary.
This behavior and the ill-posedness of the underlying optimization problem suggest that {implicit regularization} is occurring. %
In fact, %
the interpolation \eqref{eq:log_mut_f} 
can alternately be grounded in an \textit{explicit} regularization approach. Many generative models which make use of DMT can be identified with solutions of %
\textit{mean-field games} (MFGs), which are infimizations of structured cost functionals over paths of measures $\rho$ and drifts $v$ jointly satisfying a FPE \citep{zhangMeanfieldGamesLaboratory2023}. A particular MFG which fits into the framework of \citet{zhangMeanfieldGamesLaboratory2023} is 
\begin{multline}
	\inf_{v, \rho} \left \{ D_{\rm KL}(\rho(1) \| \pi) + \int_0^1 (1 - t)D_{\rm KL}(\rho(t) \| \eta) + t D_{\rm KL}(\rho(t) \| \pi) \, \rmd t + \int_0^1 \E_{\rho(t)}[L(X_t, v(X_t, t))] \, \rmd t \right \} \\ \text{s.t. } \ltfrac{\partial \rho }{\partial t} + \nabla \cdot(\rho v) = 0, \quad \rho(0) = \eta. 
	\label{eq:kfrflow_mfg}
\end{multline}
In \eqref{eq:kfrflow_mfg}, the {terminal cost} $D_{\rm KL}(\rho(1) \| \pi )$ encourages $\rho(1) \approx \pi$, while $\int_0^1 \E_{\rho(t)}[L(X_t, v(X_t, t))] \, \rmd t$ is an action cost used to penalize $v$; a typical choice is $L(x, v) = \frac{1}{2} |v |^2$. We choose \textit{interaction costs} 
$
\calI_t(\rho) = (1 - t)D_{\rm KL}(\rho \| \eta) + t D_{\rm KL}(\rho \| \pi)
$, $t \in [0,1]$, 
because they are minimized by $\mu(t) \propto \eta^{1-t}\pi^t$ \citep{amari2016information}. Thus, the solution $\rho(t)$ to \eqref{eq:kfrflow_mfg} will be close to $\mu(t)$ to the extent that it does not incur large action costs. The optimality conditions for \eqref{eq:kfrflow_mfg} imply that %
\begin{equation}
\log \rho(\cdot, t) = (1-t) \log \eta(\cdot) + t \log \pi(\cdot) - {\ltfrac{\partial U(\cdot, t)}{\partial_t} + H(\cdot, \nabla U(\cdot, t)) } - c(t) ,
\label{eq:rho_t_tilted}
\end{equation}
i.e., $\rho(t)$ is a {tilting} of the geometric mixture $\mu(t)$, similar to the model posed in \eqref{eq:log_mut_f}. \rev{In \eqref{eq:rho_t_tilted}, $U$ is the value function and $H$ is the Hamiltonian; see \cref{app:mfgs} for details.}

\section{Path identification via regularization}
Given the surprising performance of the learned interpolation approach \cite{mateLearningInterpolationsBoltzmann2023a} %
and its connection to mean-field games \citep{zhangMeanfieldGamesLaboratory2023} or related control problems, we propose to identify tilted paths of measures 
\[
\log \rho^g(x, t) = \log \rho^{\rm ref}(x, t) + g(x,t) - \log Z(t),
\]
and corresponding velocity fields $v$ for ODE transport by solving control problems of the form %
\begin{equation}
\inf_{v \in \calV, g \in \calG} \| v \|^2_\calV + \lambda_g \| g \|^2_\calG \quad
\text{s.t. } - \nabla \cdot(v \rho^g) = \rho^g (\partial_t \log \rho^g), \quad \rho^g \propto \rho^{\rm ref}e^{g}, \quad g(\cdot, 0) = g(\cdot, 1) \equiv 0.
\label{eq:control_problem_banach_space}
\end{equation}
In the above, $\rho^{\rm ref}: [0, 1] \to \calP(\R^d)$ is a reference path of measures (such as $\mu(t)$), $g: [0, 1] \times \R^d \to \R$ is a perturbation taken in a Banach space $\calG$, $Z(t) = \int_{\R^d} \rho^{\rm ref}(x, t) e^{g(x, t)} \, \rmd x$ is the normalizing constant, and $\lambda_g > 0$ is a regularization parameter. We additionally take $v: \R^d \times [0,1] \to \R^d$ in a Banach space $\calV$. 
We justify the formulation \eqref{eq:control_problem_banach_space} 
as follows:
\begin{itemize}
    \item Parametrizing $\rho^g$ as a tilting is tractable {and} expressive. Tilted measures are already used to obtain diffusion-based samplers from unnormalized densities via stochastic optimal control \citep{zhangPathIntegralSampler2022,havensAdjointSamplingHighly2025a,bernerOptimalControlPerspective2023,vargasDenoisingDiffusionSamplers2022}, and
    fine-tuning of diffusion models is frequently cast as one of sampling from a tilting of the distribution of the base model (e.g., \citep{domingo-enrichAdjointMatchingFinetuning2025}).  %

    \item  \Cref{eq:control_problem_banach_space} captures a wider range of penalties on $v$ than those that arise in MFGs \citep{zhangMeanfieldGamesLaboratory2023}. Note that if we take $\calV = L^2([0, 1], V)$ to be a Bochner space, where $V$ is an appropriate Banach space,
we obtain a penalty 
$
\| v \|_\calV^2 = \int_0^1 \| v(\cdot, t) \|^2_{V} \, \rmd t
$ akin to the action cost in a MFG \eqref{eq:kfrflow_mfg}. 
    Action costs in MFGs, however, must be of the form $\int_0^1 \E_{\rho_t}[L(X_t, v_t(X_t))] \, \rmd t$, precluding the use of, e.g., Sobolev or reproducing kernel Hilbert space (RKHS) norms to regularize $v$. %
    Recent works suggest that \textit{smoothness} plays an important role in convergence of learned DMT models \citep{beylerConvergenceDeterministicStochastic2025,tsimposOptimalSchedulingDynamic2025}, and we argue that it is important to capture this explicitly.

    \item The constraints in \eqref{eq:control_problem_banach_space} enforce $\rho^g(1) = \pi$ rather than encouraging $\rho^g(1) \approx \pi$ via a terminal cost. Our formulation may thus be better able to capture the regularization phenomena occurring in the approach of \citet{mateLearningInterpolationsBoltzmann2023a} and is also more relevant to other DMT approaches which enforce $\rho(1) = \pi$, such as stochastic interpolants \citep{albergoStochasticInterpolantsUnifying2023,liuFlowStraightFast2022a,lipmanFlowMatchingGenerative2022}. %

\end{itemize}

\subsection{Comparison to other control approaches for sampling and path identification} 
\rev{
Several works use a stochastic optimal control (SOC) approach to construct samplers (given an unnormalized density) as solutions to a Schrödinger bridge problem \citep{zhangPathIntegralSampler2022,havensAdjointSamplingHighly2025a,bernerOptimalControlPerspective2023,vargasDenoisingDiffusionSamplers2022}. 
The SOC formulation, which can also be cast as a mean-field game \cite{zhangMeanfieldGamesLaboratory2023}, is 
\begin{multline}
	\min_{u \in \calU} \E\left[ \int_0^1 \ltfrac{1}{2} \| u(X_t^u, t) \|^2  \, \rmd t + \log \ltfrac{\rho^{\rm ref}(\cdot, 1)}{\pi} (X_1^u) \right] \quad
	\text{s.t.} \quad \rmd X_t^u = \sigma(t) u(X_t^u, t )  \, \rmd t + \sigma(t) \, \rmd W_t, \\ X_0^u = 0,
	\label{eq:soc_prob_sb}
\end{multline}
where $\calU$ is a set of allowable controls, $\sigma: [0,1] \to \R^{d \times d}$ is a diffusion coefficient, and $\rho_{\rm ref}(\cdot, 1)$ is the $t = 1$ density of the uncontrolled process, e.g., 
\begin{equation}
\rmd X_t = \sigma(t) \, \rmd W_t, \quad t \in [0, 1], \quad X_0 = 0.
\label{eq:ref_process}
\end{equation} 
The motivation for adopting \eqref{eq:soc_prob_sb} is that one can show, via Girsanov's theorem, that the optimally controlled process $(X_t^{u^*})_{t \in [0,1]}$ has terminal distribution $\rho^{u^*}(\cdot, 1) = \pi$. The path measure of the process $(X_t^{u^*})_{t \in [0,1]}$ is in fact the Schrödinger bridge (SB) between $\eta = \delta_0$ and $\pi$ with base process \eqref{eq:ref_process}. %
While we also use control in our framework \eqref{eq:control_problem_banach_space}, %
we seek a path of measures resulting in \textit{smooth} dynamics, whereas the SB seeks a path of measures which is as close as possible, in KL divergence, to a reference path while satisfying desired terminal and initial conditions. When the SB problem is cast as an SOC problem, the $L_2$ norm of the drift is penalized, which promotes small magnitude but not necessarily smoothness, and the terminal condition is replaced with a terminal cost.
Our approach also differs from \eqref{eq:soc_prob_sb} in that we focus on ODEs rather than SDEs; we assume that $\eta$ has a density (i.e., is not a Dirac); and we use explicit boundary conditions to ensure $\rho^g(1) = \pi$. 
}

\rev{Another related recent work is \citet{hernandezPDPOParametricDensity2025}, which considers action-minimization problems for identifying paths between probability measures. Like our framework, \cite{hernandezPDPOParametricDensity2025} includes more general costs via an interaction energy term and enforces $\rho(0) = \eta$ and $\rho(1) = \pi$ via explicit boundary conditions. The motivation in \cite{hernandezPDPOParametricDensity2025} is to enable obstacle avoidance and to incorporate other application-specific costs in settings such as robotics, whereas our aim is principled design of DMT-based samplers. 
Numerically, \cite{hernandezPDPOParametricDensity2025} recasts the action-minimization problem as a static transport problem and lifts to a space of parametric pushforward measures, which is quite different from the dynamic PDE-constrained optimization approach we adopt here.}

\section{Numerical approach \& experiments}
\label{sec:numerics}
Here we consider \eqref{eq:control_problem_banach_space} with $\calV = \calH_v$ and $\calG = \calH_g$ as follows: $\calH_g$ is a scalar-valued RKHS \citep{berlinet2011reproducing} with kernel $K_g: Y \times Y \to \R$, where $Y = \R^d \times [0,1]$, and $\calH_v$ is a \textit{vector-valued} RKHS \citep{alvarezKernelsVectorValuedFunctions2012,kadriOperatorvaluedKernelsLearning2016}. We take $\calH_v$ to be {curl-free} and identify $v = \nabla u$, where $u$ is an element of a scalar-valued RKHS $\calH_u$ with kernel $K_u: Y \times Y \to \R$ \rev{(in our one-dimensional example this is WLOG)}. %
The problem \eqref{eq:control_problem_banach_space} is then 
\begin{multline}
\inf_{u \in \calH_u, g \in \calH_g} \| u \|^2_{\calH_u} + \lambda_g \| g \|^2_{\calH_g} \quad
\text{s.t. } - \nabla \cdot(\rho^g \nabla u ) = \rho^g (\partial_t \log \rho^g), \quad \rho^g \propto \mu e^{g}, \\ g(\cdot, 0) = g(\cdot, 1) \equiv 0.
\label{eq:control_problem_RKHS}
\end{multline}
To solve \eqref{eq:control_problem_RKHS} we %
employ the Gaussian-process PDE (GP-PDE) solution method of \cite{chenSolvingLearningNonlinear2021a}. In brief, we enforce the PDE constraint and the boundary condition $g(\cdot, 0) = g(\cdot, 1) = 0$ at finite sets of collocation points on the interior and boundary of $X$. %
Representer theorems for $u$ and $g$ \citep{Owhadi_Scovel_2019} simplify $\| g \|_{\calH_g}^2$ and $\| u \|_{\calH_u}^2$ and we relax the constraints,  ultimately obtaining the 
equivalent discrete problem  
\begin{equation}
\inf_{ \substack{\bfz_u \in \R^{(d+1)J}, \, \bfc \in\R^N \\ \bfz_g \in \R^{(d+1)J + J_b} }} 
\bfz_u\t K_u(\varphi, \varphi)\inv \bfz_u + \lambda_g \bfz_g\t K_g(\psi, \psi)\inv \bfz_g   + \lambda_{\rm pde} \sum_{j=1}^{J} \left| F_j(z^1_j, \bfz^2_j, \bfz^3_j, z^4_j, \bfc) \right|^2 + \lambda_{\rm bc} \sum_{j=1}^{J_b} | z_j^5 |^2, 
\label{eq:relaxed_problem}
\end{equation}
where $\lambda_{\rm pde}, \lambda_{\rm bc} > 0$ are regularization parameters, $\{F_j: j \in [J]\}$ encode the PDE constraint, and $\bfz_u$ and $\bfz_g$ completely parametrize the optimal $u$ and $g$. 
We use a  Levenberg-Marquardt algorithm to solve \eqref{eq:relaxed_problem} with a Cholesky change-of-variables as advocated in \cite{jalalianDataEfficientKernelMethods2025}.
As proof of concept, we use  
\eqref{eq:control_problem_RKHS} and \eqref{eq:relaxed_problem} to find a path $\rho^g$ and velocity $v_{g} = \nabla u_{g}$ for ODE transport between $\eta = \calN(0, 1)$ and $\pi = \frac{2}{3}\calN(-8, 1) + \frac{1}{3}\calN(4, 1)$, with %
$\rho^{\rm ref} = \mu$. For comparison, we use a GP-PDE approach to directly compute a velocity field $v_{\rm ref} = \nabla u_{\rm ref}$ for transport along $\rho^{\rm ref}$. %
Both approaches use the same collocation points and kernels; in particular, we take $K_u((x, t), (x', t')) = K_g((x, t), (x', t')) = K_x(x, x')K_t(t, t')$, where $K_x$ and $K_t$ are kernels on $\R$.
See \cref{app:optimal_recovery,app:experimental_details} for further details. 
\begin{figure}[h]
    \centering
    \begin{subfigure}{0.32\linewidth}
    \centering 
    \subcaption*{\small $\mu(x, t) \propto \eta(x)^{1-t} \pi(x)^t$}
        \includegraphics[width=\linewidth]{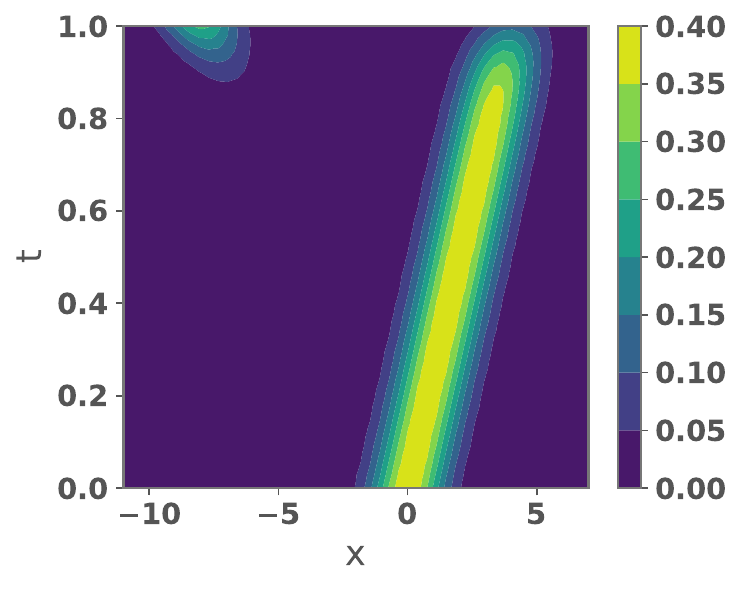}
    \end{subfigure}
\begin{subfigure}{0.32\linewidth}
\centering 
        \subcaption*{$e^{g(x, t)}$}
        \includegraphics[width=\linewidth]{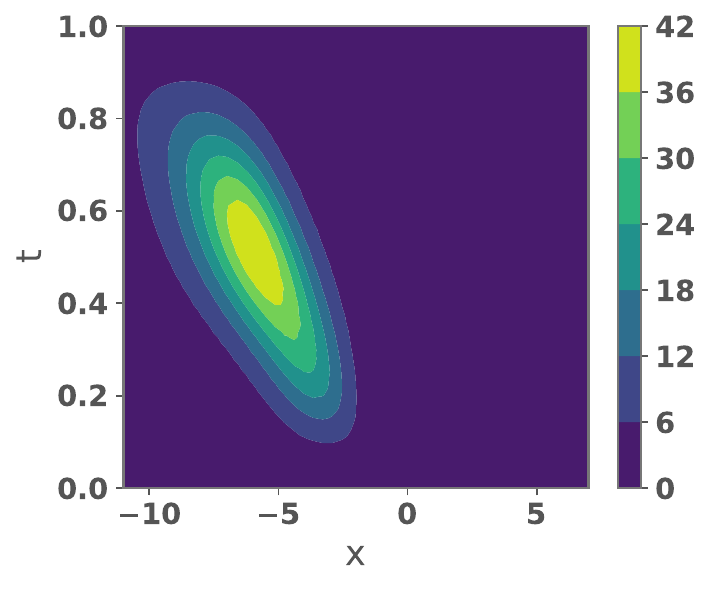}
\end{subfigure}
\begin{subfigure}{0.32\linewidth}
\centering 
    \subcaption*{\small $\rho^g(x, t) \propto \mu(x, t) e^{g(x, t)}$}
        \includegraphics[width=\linewidth]{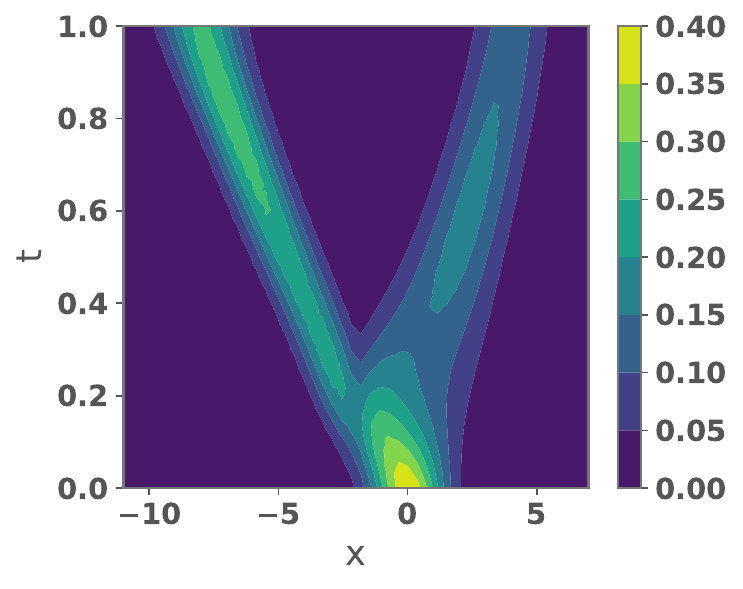}
\end{subfigure}
    \caption{Space-time plots of the reference path $\mu(x, t) \propto \eta(x)^{1-t}\rho(x)^t$ (left), the tilting $e^{g(x, t)}$ (center), and the path $\rho^g(x, t) \propto \mu(x, t)e^{g(x, t)}$ resulting from \eqref{eq:relaxed_problem} (right).} %
    \label{fig:spacetime-plots-density}
\end{figure}

In \cref{fig:learned_interp_1D,fig:spacetime-plots-density} we show the two paths, $\rho^{\rm ref} = \mu$ and $\rho^g$, and in \cref{fig:learned_interp_1D} we show samples generated using the corresponding velocities, $v_{\rm ref}$ and $v_{g}$. %
The tilting $e^g$ recovered from \eqref{eq:relaxed_problem} eliminates the teleportation present in $\mu$, leading to better-quality samples generated by $v_g$. %
\rev{In \cref{fig:trajectories} we display the trajectories of particles sampled from $\eta$ and transported by $v_{\rm ref}$, $v_{g}$, and the velocity corresponding to the McCann interpolant \cite{mccannConvexityPrincipleInteracting1997} (computed analytically in this 1D example). We see that, in addition to placing more samples in the left mode of $\pi$ than $v_{\rm ref}$, the learned velocity $\nabla u_g$ is spatially smoother than the McCann velocity. This result is similar in flavor to that of \citet[Figure 3]{tsimposOptimalSchedulingDynamic2025}, wherein a time-rescaling is applied to the McCann interpolant to obtain a smoother velocity field. Our approach differs from \cite{tsimposOptimalSchedulingDynamic2025} in that we do not use the McCann interpolant as a starting point and that the path of densities itself, rather than just the schedule, is allowed to deviate from the reference.} 
In \cref{fig:spatial_RKHS_norms} we plot the spatial RKHS norms $\| u_g(\cdot, t) \|_{\calH_x}$ and $\| u_{\rm ref}(\cdot, t) \|_{\calH_x}$, where $\calH_x$ is the RKHS with kernel $K_x$, as a function of $t$. %
We see that $\| u_{\rm ref}(\cdot, t) \|_{\calH_x}$ increases by more than tenfold over the course of $[0, 1]$ in order to capture the teleportation in $\rho^{\rm ref}$, while $\| u_g \|_{\calH_x}$ stays relatively constant. %
We assess the quality of the samples generated by $v_g$ and $v_{\rm ref}$ in \cref{tab:sample_metrics}. While $v_g$ does not sample perfectly, %
it still represents a dramatic improvement over $v_{\rm ref}$.

\section{Conclusion}
We have presented a flexible, general control framework \eqref{eq:control_problem_banach_space} for identifying paths of measures for DMT as tiltings of accessible reference paths. Our framework enables the promotion of {smoothness} of the associated dynamics via penalization with, e.g., Sobolev or RKHS norms, and can serve as the basis for a range of numerical implementations. We have used one such implementation to generate proof-of-concept results demonstrating clear benefits of using a learned path with smooth dynamics, but looking ahead we are considering other formulations based on alternate functional penalties, for example, Bochner space norms on $v$ and $g$. We anticipate that our framework will enable us to discrern the relative roles of spatial and temporal regularity in influencing the tractability of a given path, and ultimately inform better choices of path in sampling applications, like Bayesian inference and data assimilation, where annealing is often employed and the reference $\eta$ cannot be modified. %
\clearpage 
\begin{ack}
AM and YM were supported by the Office of Naval Research, SIMDA (Sea Ice Modeling and Data Assimilation) MURI, award number N00014-20-1-2595 (Dr.~Reza Malek-Madani and Dr.~Scott Harper). AM was additionally supported by the NSF Graduate Research Fellowship under Grant No.\ 1745302 and by Google through an MIT Schwarzman College of Computing Future Research Cohort Fellowship. 
BH was supported by the National Science Foundation 
grants DMS-233767 (CAREER: Gaussian Processes for Scientific Machine
Learning: Theoretical Analysis and Computational Algorithms), 
and DMS-2208535 (Machine Learning for Bayesian Inverse Problems).
AM thanks Jannis Chemseddine, Markos Katsoulakis, Gabriele Steidl, and Benjamin Zhang for helpful discussions and thanks Alexander Hsu and Juan Felipe Osorio Ramierez for help with the GP-PDE implementation. 
\end{ack}

\appendix 
\section{Additional numerical results}
Here we provide additional figures and tables corresponding to the experiment in \cref{sec:numerics}.

\begin{figure}[h] 
\centering 
\begin{subfigure}{0.32\linewidth}
\centering 
    \subcaption*{Reference Trajectories}
   \includegraphics[width=\linewidth]{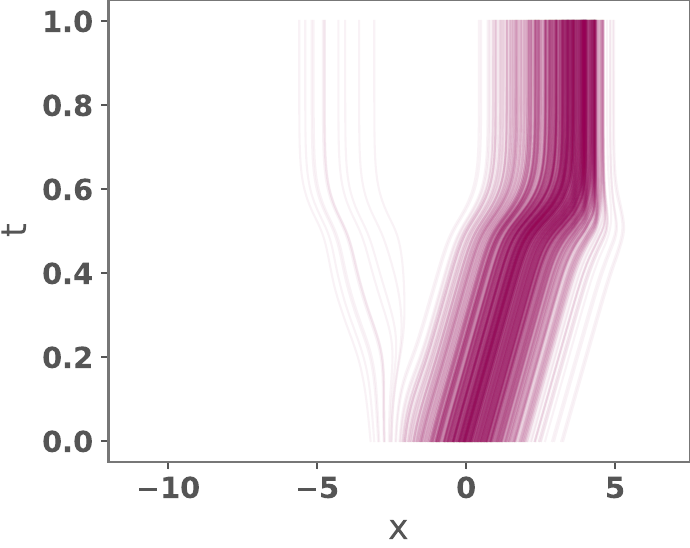}
\end{subfigure}
\begin{subfigure}{0.32\linewidth}
\centering 
\subcaption*{Learned Trajectories}
    \includegraphics[width=\linewidth]{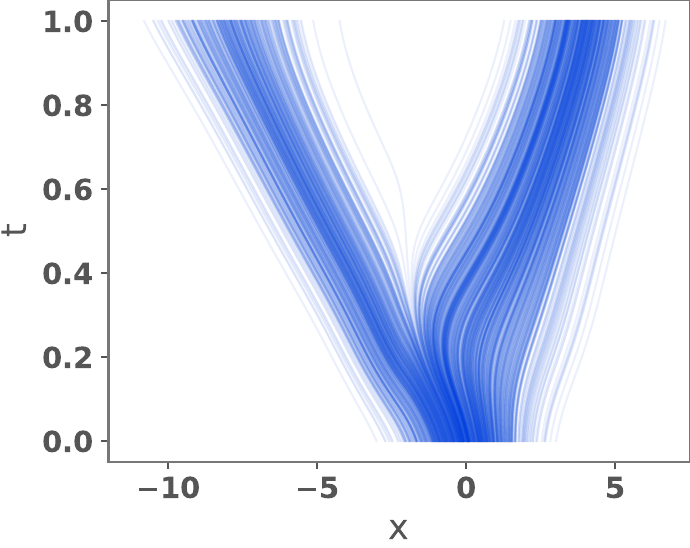}
\end{subfigure}
\begin{subfigure}{0.32\linewidth}
\centering 
    \subcaption*{McCann Trajectories}
    \includegraphics[width=\linewidth]{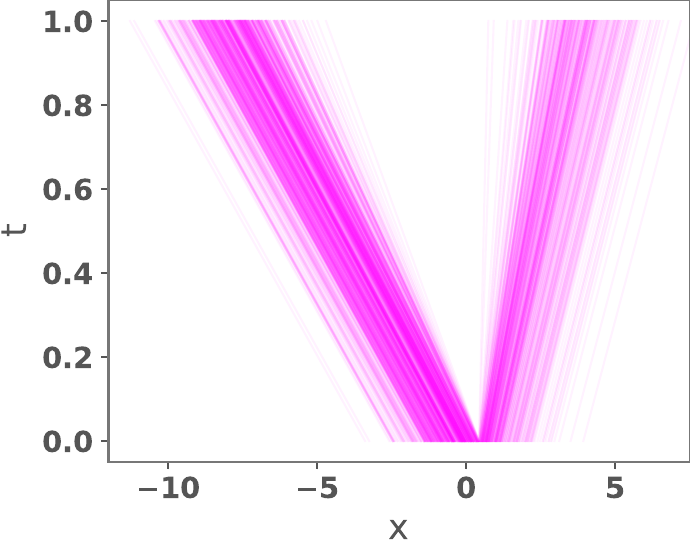}
\end{subfigure}
\caption{Trajectories corresponding to three different velocity fields for DMT between $\eta$ and $\pi$: the reference velocity $v_{\rm ref} = \nabla u_{\rm ref}$ (left), the learned velocity $v_g = \nabla u_{g}$ (center), and the McCann interpolant velocity (right).  The learned velocity $v_{g}$ places more mass in the left mode than the reference velocity $v_{\rm ref}$ and is {spatially smoother} than the McCann interpolant velocity.}
\label{fig:trajectories}
\end{figure} %

\begin{figure}[h]
\centering 
    \begin{subfigure}{0.24\linewidth}
    \centering 
    \subcaption*{$u_{\rm ref}(x, t)$}
        \includegraphics[width=\linewidth]{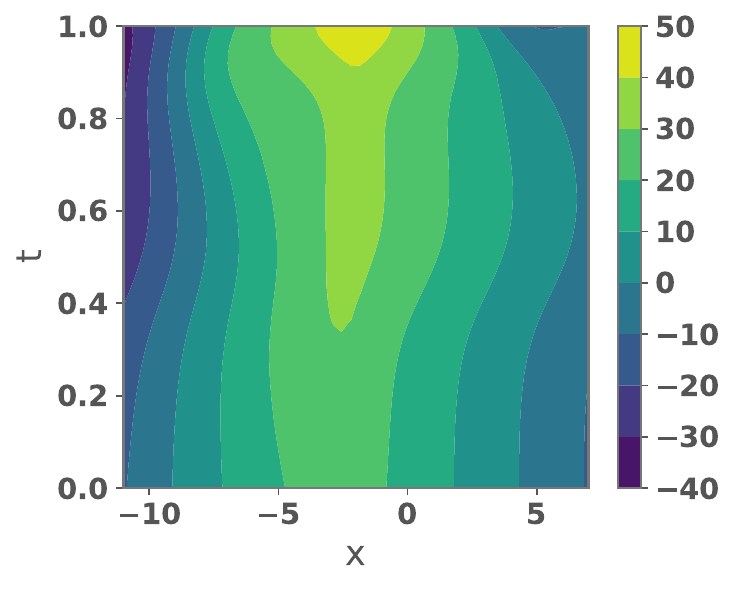}
    \end{subfigure}
    \begin{subfigure}{0.24\linewidth}
    \centering
        \subcaption*{$ u_{g}(x, t)$}
        \includegraphics[width=\linewidth]{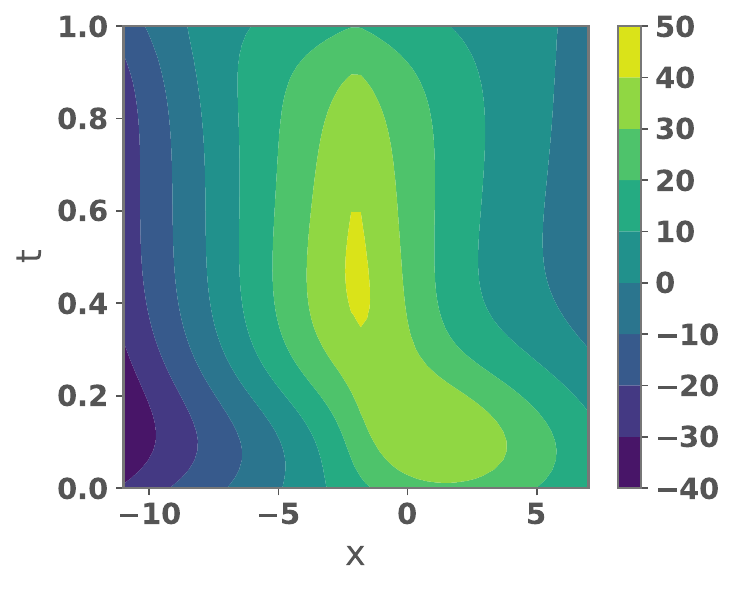}
    \end{subfigure}
    \begin{subfigure}{0.24\linewidth}
    \centering
        \subcaption*{$ \mu(x, t) u_{\rm ref}(x, t) $}
        \includegraphics[width=\linewidth]{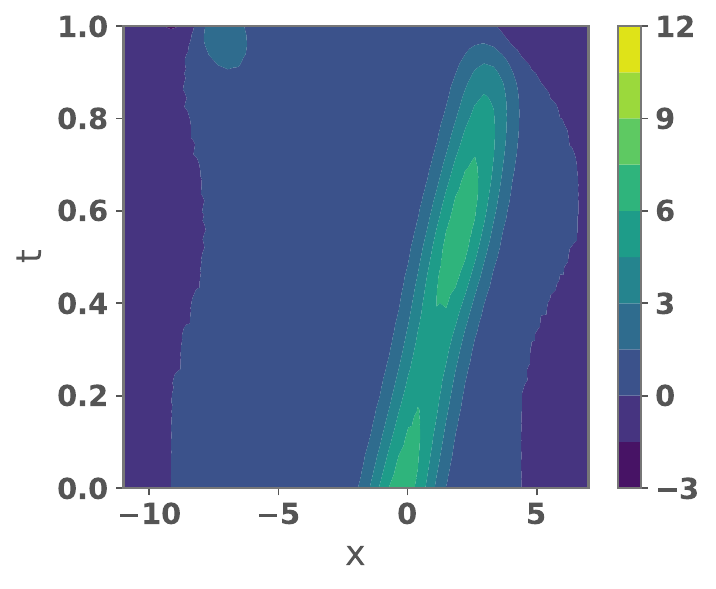}
    \end{subfigure}
    \begin{subfigure}{0.24\linewidth}
    \centering
    \subcaption*{$\rho^g(x, t) u_{g}(x, t) $}
        \includegraphics[width=\linewidth]{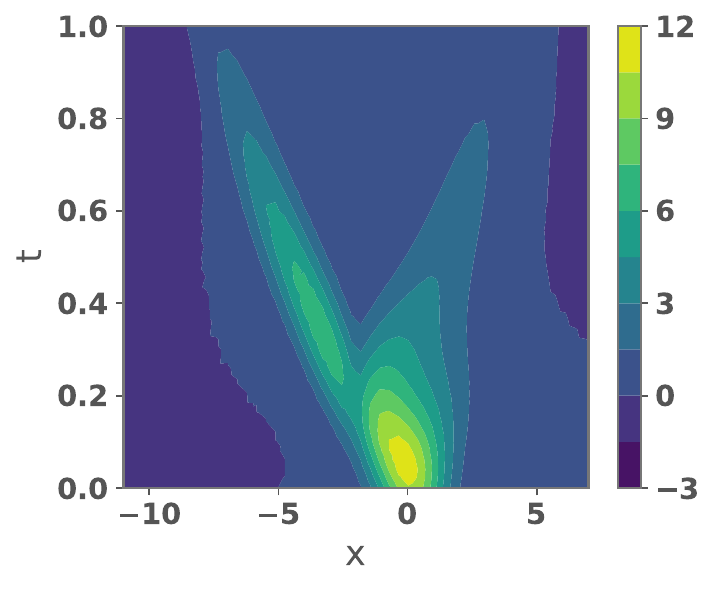}
    \end{subfigure}
    \begin{subfigure}{0.24\linewidth}
    \centering 
    \subcaption*{$\nabla u_{\rm ref}(x, t)$}
        \includegraphics[width=\linewidth]{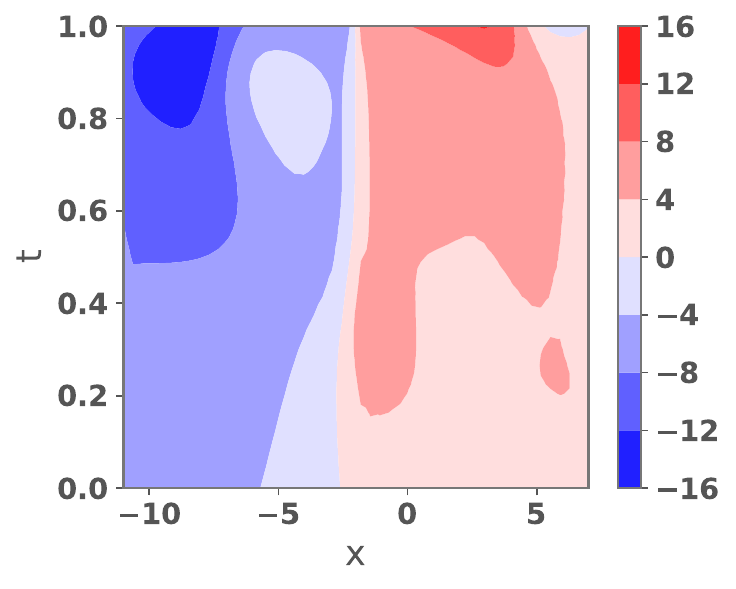}
    \end{subfigure}
    \begin{subfigure}{0.24\linewidth}
    \centering
        \subcaption*{$\nabla u_{g}(x, t)$}
        \includegraphics[width=\linewidth]{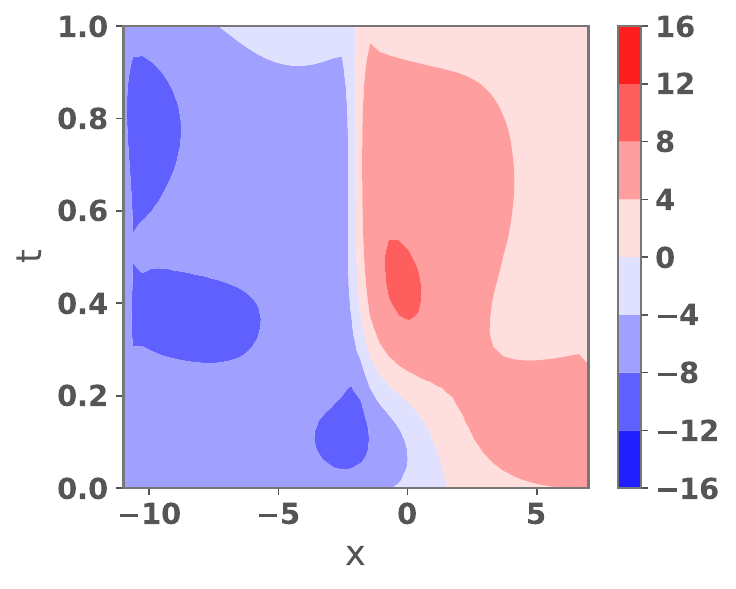}
    \end{subfigure}
    \begin{subfigure}{0.24\linewidth}
    \centering
        \subcaption*{$ \mu(x, t) \nabla u_{\rm ref}(x, t) $}
        \includegraphics[width=\linewidth]{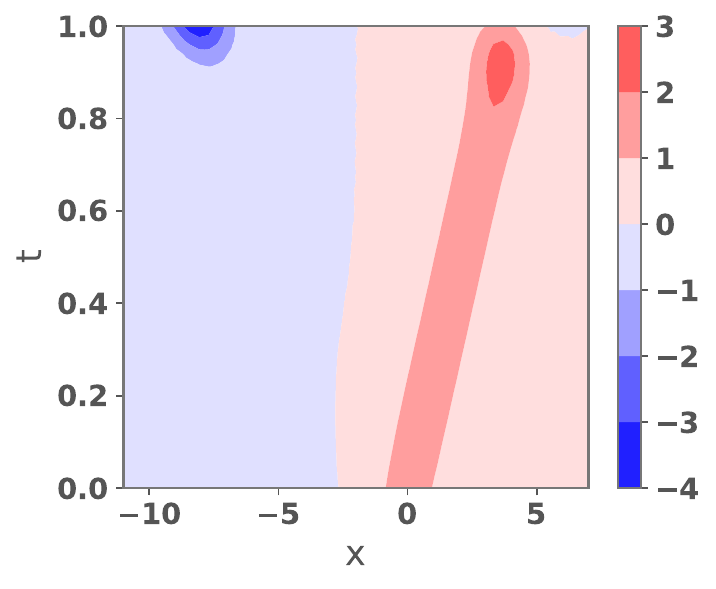}
    \end{subfigure}
    \begin{subfigure}{0.24\linewidth}
    \centering
    \subcaption*{$\rho^g(x, t) \nabla u_{g}(x, t) $}
        \includegraphics[width=\linewidth]{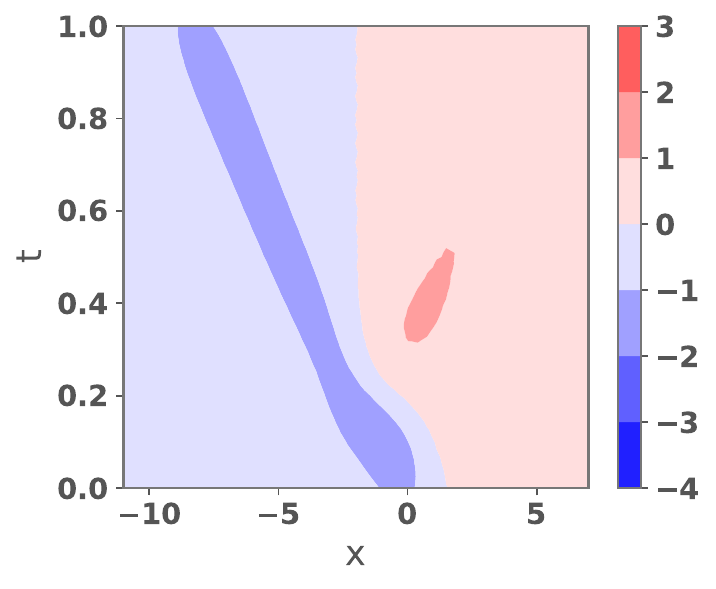}
    \end{subfigure}
    \caption{Potentials $u_{\rm ref}$ and $u_g$ and velocity fields $v_{\rm ref} = \nabla u_{\rm ref}$ and $v_{g} = \nabla u_g$ corresponding to the geometric path $\rho^{\rm ref} = \mu$ and the path $\rho^g$ obtained from \eqref{eq:relaxed_problem}. In the first two columns of panels we show the absolute potentials/velocities, and in the second two columns we show the potentials/velocities weighted by their respective probability densities, which better capture how the mass is moving.}
        \label{fig:velocities_potentials}
\end{figure}

\begin{figure}[h]
\centering 
    \includegraphics[width=0.4\linewidth]{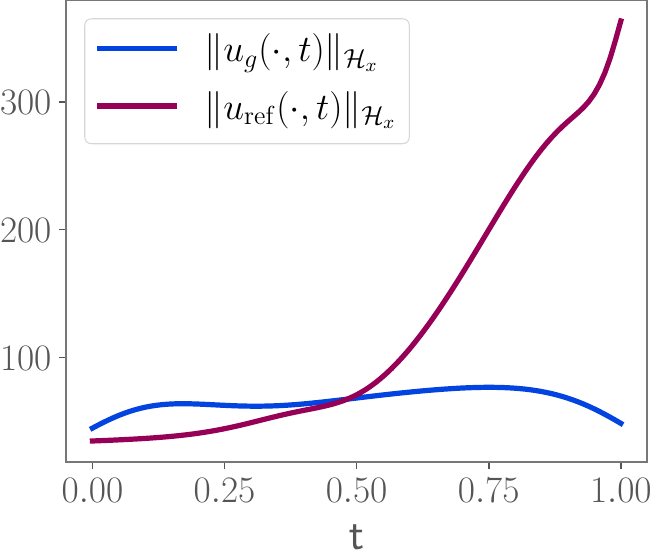}
    \caption{Spatial RKHS norms of $u_g(\cdot, t)$ (blue) and $u_{\rm ref}(\cdot, t)$ (red) as a function of time. }
\label{fig:spatial_RKHS_norms}
\end{figure}

\begin{table}[h]
    \centering
    \begin{tabular}{cp{1.5cm}p{2cm}p{2cm}cc}
         &  Fraction in left mode & Relative Error in Mean $\downarrow$ & Relative Error in Variance $\downarrow$ & MMD $\downarrow$ & $\| u \|_{\calH}$ \\ \toprule
        Reference Interpolation &  0.005 & 1.80 & 0.96 & 0.743 & 770. \\ 
        Learned Interpolation &  0.375 & 0.88 & 0.016 & 0.137 & 136 \\ 
        \textit{Ground Truth Samples} & \textit{0.654} & \textit{0.040} & \textit{0.024} & \textit{$7.21 \times 10^{-4}$} & \textit{n/a} 
    \end{tabular}
    \vskip 0.25cm
    \caption{Quality metrics evaluated on 1000 samples generated by $v_g$ and by $v_{\rm ref}$. We evaluate the same metrics on 1000 ground-truth samples from $\pi$ for comparison. In truth $2/3$ of the mass of $\pi$ belongs in the left mode, the mean of $\pi$ is $-4$ and the variance of $\pi$ is 33.}
    \label{tab:sample_metrics}
\end{table}

\section{Optimality conditions for mean-field game}
\label{app:mfgs}
In \cref{sec:fix_and_mfg} we introduce the mean-field game 
\begin{multline}
	\inf_{v, \rho} \left \{ D_{\rm KL}(\rho(1) \| \pi) + \int_0^1 (1 - t)D_{\rm KL}(\rho(t) \| \eta) + t D_{\rm KL}(\rho(t) \| \pi) \, \rmd t + \int_0^1 \E_{\rho(t)}[L(X_t, v(X_t, t))] \, \rmd t \right \} \\ \text{s.t. } \ltfrac{\partial \rho }{\partial t} + \nabla \cdot(\rho v) = 0, \quad \rho(0) = \eta. 
	\label{eq:kfrflow_mfg_app}
\end{multline}

The optimality conditions for this game consist of a coupled system of 
a Hamilton-Jacobi-Bellman equation \eqref{eq:HJB} and a continuity equation \label{eq:FP},
\begin{align}  
		-\sdfrac{\partial U(x, t)}{\partial t} + {H(x, \nabla U(x,t))} = {\log \rho(x, t) - (1-t)\log \eta(x) - t \log \pi(x) } + c(t) \label{eq:HJB} \\
		\sdfrac{\partial \rho(x,t)}{\partial_t} - \nabla \cdot(\rho(x,t) {\nabla_2 H(x, \nabla U(x,t))}) = 0  \label{eq:FP} \\
		U(x, 1) = {b + \log \sdfrac{\rho(\cdot, 1)}{\pi}(x)}, \quad  \rho(\cdot, 0) = \eta, \label{eq:TC} %
\end{align}
where $H(x, p) = \sup_v [-p \t v - L(x, v)]$ is the Hamiltonian\footnote{e.g., if $L(x, v) = \frac{1}{2}| v |^2$, then $H(x, p) = \frac{1}{2}| p |^2$ and $\nabla_2H(x, p) = p$.}, $U: \R^d \times [0,1] \to \R$ is the value function, $b \in \R$ is constant, and $c: [0, 1] \to \R$ is a time-varying constant. \Cref{eq:HJB,eq:FP,eq:TC} follow from standard results in control theory \cite{lasryMeanFieldGames2007,zhangMeanfieldGamesLaboratory2023}.

\section{GP-PDE computational approach}
\label{app:optimal_recovery}
\subsection{For the reference solution}
\label{app:linear_PDEs}
Before describing the GP-PDE solution approach to \eqref{eq:control_problem_RKHS}, 
we first describe the kernel collocation approach used to approximately solve the elliptic equation 
\begin{equation}
- \nabla \cdot(\mu(x, t) \nabla u_{\rm ref}(x, t)) = \mu(x, t)\left(\log\ltfrac{\eta}{\pi}(x) - \E_{\mu(\cdot, t)}[\log \ltfrac{\eta}{\pi}] \right ),
\label{eq:elliptic_pde}
\end{equation}
which recovers a velocity field $\nabla u_{\rm ref}$ for transport along the geometric mixture $\mu(t) = \eta^{1 - t} \pi^t$. This approach is used as a basis for comparison to \eqref{eq:control_problem_RKHS} and is a {building block} of the approach to \eqref{eq:control_problem_RKHS}. 

We denote the linear operator on the LHS of \eqref{eq:elliptic_pde} by $\calL u \coloneqq - \nabla \cdot (\mu \nabla u)$, and denote the right-hand-side of \eqref{eq:elliptic_pde} by $f(x, t) \coloneqq \mu(x, t)(\log\frac{\eta}{\pi}(x) - \E_{\mu(\cdot, t)}[\log \frac{\eta}{\pi}] )$. Thus, the PDE \eqref{eq:elliptic_pde} reads $\calL u_{\rm ref} = f$. 

We only enforce the PDE \eqref{eq:elliptic_pde} at a set of collocation points $\{ (x_j, t_j) \}_{j=1}^{J} \subseteq Y$, obtaining 
\begin{equation}
(\calL u_{\rm ref})(x_j, t_j) = f(x_j, t_j), \quad j = 1, \dots, J.
\label{eq:colloc_linear_pde}
\end{equation}
We now seek a solution $u_{\rm ref}: Y \to \R$ to \eqref{eq:colloc_linear_pde}, where $Y = \R^d \times [0,1]$, in the RKHS $\calH_u$ with kernel $K_u$ having \textit{minimum norm},
\begin{equation}
u_{\rm ref} =  
\argmin_{u \in \calH_u} \| u \|_{\calH_u}^2 \quad \text{s.t.  } (\calL u)(x_j, t_j) = f(x_j, t_j), \quad j = 1, \dots, J.
\label{eq:or_problem_linear_pde}
\end{equation}
For $j = 1, \dots, J$, let $\phi_j: \calH_u \to \R$ denote the linear functional 
\[
\phi_j(u) = (\calL u)(x_j, t_j), 
\]
and let $\phi = (\phi_1, \dots, \phi_{\rm J}): \calH_u \to \R^{J}$ be the linear feature map comprised of $\phi_1, \dots, \phi_{J}$. Denoting $\bff = (f(x_1, t_1), \dots, f(x_J, t_J)) \in \R^{J}$, the problem \eqref{eq:or_problem_linear_pde} reads
\begin{equation}
u_{\rm ref} =  
\argmin_{u \in \calH_u} \| u \|_{\calH_u}^2 \quad \text{s.t.  } \phi(u) = \bff. 
\label{eq:or_problem_vectorized}
\end{equation}
\cref{eq:or_problem_vectorized} is an \textit{optimal recovery problem} and has a well-known solution arising from representer theorems on RKHS (e.g., \citet{Owhadi_Scovel_2019}, see also \citet{chenSolvingLearningNonlinear2021a,jalalianDataEfficientKernelMethods2025}), namely
\begin{equation}
u_{\rm ref}(\cdot) = K_u(\cdot, \phi) K_u(\phi, \phi)\inv \bff.
\label{eq:or_soln_ref}
\end{equation}
In \eqref{eq:or_soln_ref} $K_u: Y \to \R^{1 \times J}$ is a vector field with elements 
\begin{equation}
K_u(\cdot, \phi) = \begin{pmatrix}
    K_u(\cdot, \phi)_1 & \cdots & K_u(\cdot, \phi)_{J}
\end{pmatrix}, \quad K_u(y, \phi)_i = \phi_i(K_u(y, \cdot) ), \quad i = 1, \dots, J, 
\label{eq:vector_field_ref}
\end{equation}
and $K_u(\phi, \phi) \in \R^{J \times J}$ is a symmetric matrix with entries 
\begin{equation}
K_u(\phi, \phi)_{ij} = \phi_i(K_u(\cdot, \phi)_j), \quad i, j \in \{1, \dots, J \}.  
\label{eq:matrix_ref}
\end{equation}
Moreover, the RKHS norm of the optimal recovery solution $u^*$ is 
\[
\| u_{\rm ref} \|^2_{\calH_u} = \bff\t K_u(\phi, \phi)\inv \bff. 
\]
In our experiments, we approximate the unknown expectations $\E_{\mu(t)}[\log \frac{\pi}{\eta}]$ appearing in \eqref{eq:elliptic_pde} using quadrature, since our examples are one-dimensional. 

\subsection{For the control problem}
\label{app:control_or}
Now we return to the problem in \cref{eq:control_problem_RKHS},
\begin{equation}
\inf_{u \in \calH_u, g \in \calH_g} \| u \|^2_{\calH_u} + \lambda \| g \|^2_{\calH_g} \quad
\text{s.t. } - \nabla \cdot(\rho^g \nabla u ) = \rho^g (\partial_t \log \rho^g), \quad \rho^g \propto \mu e^{g}, \quad g(\cdot, 0) = g(\cdot, 1) \equiv 0.
\label{eq:control_problem_RKHS_app}
\end{equation}
Recall that $\calH_g$ is a scalar-valued RKHS with kernel $K_g: Y \times Y \to \R$ and $\calH_u$ is a scalar-valued RKHS with kernel $K_u: Y \times Y \to \R$. This problem is similar to \eqref{eq:or_problem_linear_pde} except that the constraint is {nonlinear} in $u$ and $g$ jointly. 
We proceed similarly as before, only enforcing the PDE constraint, which can be equivalently written 
\begin{multline}
F(x, t; g, u) \equiv 
\log \ltfrac{\pi}{\eta}(x) + \partial_t g(x, t) - \E_{\xi \sim \rho^g(t)}\left[ \log \ltfrac{\pi}{\eta}(\xi) + \partial_t g(\xi, t) \right] \\ 
- \left\langle (1 - t) \nabla \log \eta(x) + t \nabla \log \pi(x) + \nabla g(x, t), \nabla u(x, t) \right\rangle - \Delta u(x, t) = 0,
\label{eq:F}
\end{multline}
at the same finite set of points $\{ (x_j, t_j) \}_{j=1}^{J} \subseteq \R^d \times [0,1]$ used for the reference method. Likewise, we enforce the boundary conditions $g(\cdot, 0) = g(\cdot, 1) \equiv 0$ at finite sets of points on the boundary, $\{(x^0_j, 0)\}_{j=1}^{J_0}$ and $\{(x^1_j, 1)\}_{j=1}^{J_1}$, obtaining 
\begin{equation}
\inf_{u \in \calH_u, g \in \calH_g} \| u \|^2_{\calH_u} + \lambda \| g \|^2_{\calH_g} \quad
\text{s.t.} \quad \begin{cases}
    F(x_j, t_j; g, u) = 0 & j \in \{1, \dots, J\} \\ 
     g(x^0_j, 0) = 0 & j \in \{1, \dots, J_0\} \\ 
     g(x^1_j, 0) = 0 & j \in \{1, \dots, J_1\},
\end{cases}
\label{eq:control_problem_colloc_app}
\end{equation}
The first set of constraints in \eqref{eq:control_problem_colloc_app} can be expanded 
\begin{multline}
F(x_j, t_j; g, u) = \log \ltfrac{\pi}{\eta}(x_j) + \partial_t g(x_j, t_j) - C(t_j) \\ %
- \left\langle (1 - t_j) \nabla \log \eta(x_j) + t_j \nabla \log \pi(x_j) + \nabla g(x_j, t_j), \nabla u(x_j, t_j) \right\rangle - \Delta u(x_j, t_j) = 0, \\ j = 1, \dots, J.,
\label{eq:expanded_constraints}
\end{multline}
where 
\[
C(t_j) \coloneqq \E_{\xi \sim \rho^g(t_j)}\left[ \log \ltfrac{\pi}{\eta}(\xi) + \partial_t g(\xi, t_j) \right].
\]
$C(t)$ is the time-derivative of the log normalizing constant of $\rho^g$ and is typically unknown; in our implementation we learn needed evaluations of $C$ (at all distinct $t_j$ in our collocation point set), which we denote by $\bfc \in\R^N$, simultaneously with $u$ and $g$; see also \citet[Lemma 1]{mateLearningInterpolationsBoltzmann2023a}. 

Notice that the constraints \eqref{eq:expanded_constraints} only depend on the values of $\partial_t g$, $\nabla g$, $\nabla u$, and $\Delta u$ at $\{(x_j, t_j)\}_{j=1}^{J}$. Likewise, the boundary constraints in \eqref{eq:control_problem_colloc_app} only depend on the values of $g$ at $\{(x^0_j, 0)\}_{j=1}^{J_0}  \cup \{(x^1_j, 1)\}_{j=1}^{J_1}  \equiv \{(x^b_j, t^b_j)\}_{j=1}^{J_b}$, where $J_b = J_0 + J_1$. 
As such we denote these values 
\begin{equation}
\begin{rcases}
\partial_t g(x_j, t_j) &\coloneqq z_j^1 \in \R \\ 
\nabla g(x_j, t_j)  &\coloneqq \bfz_j^2 \in \R^d  \\
\nabla u(x_j, t_j) &\coloneqq \bfz_j^3 \in \R^d \\ 
\Delta u(x_j, t_j)   &\coloneqq z_j^4 \in \R, 
\end{rcases}, \quad  j = 1, \dots, J
\end{equation}
and
\begin{align*}
g(x^b_j, t^b_j) &\coloneqq z_j^5 \in \R, \quad j = 1, \dots J_b.
\end{align*}
For brevity, we also introduce notation for the known quantities in \eqref{eq:control_problem_colloc_app},
\begin{align*}
\log \ltfrac{\pi}{\eta} ( x_j ) &\coloneqq \ell_j \in \R \\ 
(1-t_j) \nabla \log \eta ( x_j ) + t_j \nabla \log \pi( x_j ) &\coloneqq \bfs_j \in \R^d , \quad j = 1, \dots J.
\end{align*}

With this notation in hand, the collocation \cref{eq:expanded_constraints,eq:control_problem_colloc_app} can be written 
\begin{equation}
F_j(z^1_j, \bfz^2_j, \bfz^3_j, z^4_j, \bfc) \equiv \ell_j + z_j^1 - C(t_j)  - \left\langle \bfs_j + \bfz_j^2 ,\, \bfz_j^3 \right\rangle  - z_j^4 = 0, \quad 
j \in \{1, \dots, J\}, 
\label{eq:colloc_concise}
\end{equation}
and 
\begin{equation}
z_j^5  = 0, \quad j \in \{1, \dots, J_b\}.
\label{eq:bcg_concise}
\end{equation}

Thus, fulfilling the constraints of \eqref{eq:control_problem_colloc_app} consists in identifying suitable values of $z^1_j$, $\bfz_j^2$, $\bfz_j^3$, and $z_j^4$, $j \in \{1, \dots, J\}$, $z_j^5$, $j \in \{1, \dots, J_b\}$, and $\bfc \in \R^N$.  
Therefore we return to \eqref{eq:control_problem_colloc_app}, replacing the constraints with the collocation \cref{eq:bcg_concise,eq:colloc_concise} and obtaining a \textit{bilevel} optimization problem
\begin{multline}
\inf_{ \substack{z^1_j, \bfz^2_j, \bfz^3_j, z_j^4, \, j \in [J]\\  z_j^5, \, j \in [J_b] \\ \bfc \in \R^N }} \left\{
\inf_{u \in \calH_u, g \in \calH_g} \| u \|^2_{\calH_u} + \lambda \| g \|^2_{\calH_g} \quad
\text{s.t.} \quad \begin{cases}
\partial_t g(x_j, t_j) &= z_j^1 \in \R, \; j \in [J] \\ 
\nabla g(x_j, t_j)  &= \bfz_j^2 \in \R^d,  \; j \in [J] \\
\nabla u(x_j, t_j) &= \bfz_j^3 \in \R^d, \; j \in [J] \\ 
\Delta u(x_j, t_j)   &= z_j^4 \in \R, \; j \in [J] \\ 
g(x^b_j, t^b_j) &= z_j^5 \in \R, \; j \in [J_b]
\end{cases} \right\}  \\[0.3cm] 
\text{s.t. } F_j(z^1_j, \bfz^2_j, \bfz^3_j, z^4_j, \bfc) = 0, \; 
j \in [J], \quad z_j^5  = 0, \quad j \in [J_b]. 
\label{eq:bilevel_opt}
\end{multline}

The inner problem in \eqref{eq:bilevel_opt}, being separable in $u$ and $g$, has a solution analogous to \eqref{eq:or_soln_ref},
\begin{equation}
u^*(\cdot) = K_u(\cdot, \varphi) K_u(\varphi, \varphi)\inv \bfz_u, \quad g^*(\cdot) = K_g(\cdot, \psi) K_u(\psi, \psi)\inv \bfz_g.
\label{eq:ustar}
\end{equation}
We use $\bfz_u \in \R^{J(d+1)}$ to denote 
\begin{equation}
\bfz_u = \begin{pmatrix}z_1^4 & \cdots & z_{J}^4 & (\bfz_1^3)\t & \cdots & (\bfz_{J}^3)\t \end{pmatrix}\t,
\label{eq:zu}
\end{equation}
and $\bfz_g \in \R^{J(d+1) + J_b}$ to denote 
\[
\bfz_g = \begin{pmatrix}z_1^1 & \cdots & z_{J}^1 & z^5_1 & \cdots & z^5_{J_b} & (\bfz_1^2)\t & \cdots & (\bfz_{J}^2)\t \end{pmatrix}\t.
\]
In \eqref{eq:ustar} $\varphi: \calH_u \to \R^{J(d+1)}$ is the linear feature map 
\begin{equation}
\begin{aligned}
\varphi(\cdot) &= \begin{pmatrix} \varphi^1(\cdot) & \cdots & \varphi^{J}(\cdot) & \varphi^{11}(\cdot) & \cdots & \varphi^{1d}(\cdot) & \cdots \cdots & \varphi^{{J}1}(\cdot) & \cdots & \varphi^{{J}d}(\cdot) \end{pmatrix}\t \\
&\equiv \begin{pmatrix} \varphi^1(\cdot) & \cdots & \varphi^{J}(\cdot) & \varphi^{{J}+1}(\cdot) &\cdots& \cdots& \cdots &\cdots&\cdots& \varphi^{{J}(d+1)}(\cdot) \end{pmatrix}\t, 
\label{eq:features_u}
\end{aligned}
\end{equation}
where the component linear functionals 
\begin{equation}
    \varphi^i(u) = \Delta u(x_i, t_i), \quad 
    \varphi^{ij}(u) = (\nabla u (x_i, t_i))_j, \quad i \in \{1, \dots, J\}, \; j \in \{1, \dots, d\},
    \label{eq:phi_functionals}
\end{equation}
give rise to the elements of $\bfz_u$. Similarly, $\psi: H_g \to \R^{J(d+1) + J_b}$ is the linear feature map 
\begin{equation}
\begin{aligned}
\psi &= \left( \psi^1,\, \dots,\, \psi^{J},\, \psi^{\rm J + 1},\, \dots,\, \psi^{J + J_b},\, \psi^{11},\,  \dots ,\, \psi^{1d} ,\, \dots ,\, \psi^{{J}1} ,\, \dots ,\, \psi^{{J}d} \right) \t \\
&\equiv \begin{pmatrix} \psi^1 & \dots & \psi^{J + J_b} & \psi^{J + J_b + 1} &\dots& \dots& \dots &\dots&\dots& \psi^{(d+1)J + J_b} \end{pmatrix}\t,\, 
\label{eq:features_u}
\end{aligned}
\end{equation}
where the component linear functionals 
\begin{gather*}
    \psi^i(g) = \partial_t g(x_i, t_i), \quad 
    \psi^{ij}(u) = (\nabla g (x_i, t_i))_j, \quad i \in \{1, \dots, J\}, \; j \in \{1, \dots, d\} \\[0.1cm] 
    \psi^{J + i}(g) = g(x^b_i, t^b_i), \quad i \in \{1, \dots, J_b\}.
    \label{eq:psi_functionals}
\end{gather*}
give rise to the elements of $\bfz_g$.

The vector fields $K_u: Y \to \R^{1 \times J(d+1)}$ and $K_g(\cdot, \psi): Y \to \R^{1 \times J(d+1) + J_b}$ are defined analogously to \eqref{eq:vector_field_ref}, and 
the symmetric matrices $K(\varphi, \varphi) \in \R^{J(d+1) \times J(d+1)}$ and $K_g(\psi, \psi) \in \R^{(J(d+1) + J_b) \times (J(d+1) + J_b)} $  are defined analogously to \eqref{eq:matrix_ref}.

The norm of $u^*$ in \eqref{eq:ustar} is $\| u^* \|_{\calH_u}^2 = \bfz_u\t K_u(\varphi, \varphi)\inv \bfz_u$ and the norm of $g^*$ is $\| g^* \|_{\calH_g}^2 = \bfz_g\t K_g(\psi, \psi)\inv \bfz_g$. These norms define the optimal value of the inner problem in \eqref{eq:bilevel_opt} such that the problem reduces to 
\begin{multline}
\inf_{ \substack{\bfz_u \in \R^{(d+1)J} \\  \bfz_g \in \R^{(d+1)J + J_b} \\ \bfc \in\R^N }} 
\bfz_u\t K_u(\varphi, \varphi)\inv \bfz_u + \lambda \bfz_g\t K_g(\psi, \psi)\inv \bfz_g    \\[-0.5cm]
\text{s.t. } F_j(z^1_j, \bfz^2_j, \bfz^3_j, z^4_j, \bfc) = 0, \; 
j \in [J], \quad z_j^5  = 0, \quad j \in [J_b]. 
\label{eq:final_constrained_problem}
\end{multline}

Following the relaxation approach of \citet{chenSolvingLearningNonlinear2021a}, in practice we exchange the constrained problem \eqref{eq:final_constrained_problem} for the penalized unconstrained problem 
\begin{equation}
\inf_{ \substack{\bfz_u \in \R^{(d+1)J} \\ \bfz_g \in \R^{(d+1)J + J_b} \\  \bfc \in\R^N }} 
\bfz_u\t K_u(\varphi, \varphi)\inv \bfz_u + \lambda \bfz_g\t K_g(\psi, \psi)\inv \bfz_g   + \lambda_{\rm pde} \sum_{j=1}^{J} \left| F_j(z^1_j, \bfz^2_j, \bfz^3_j, z^4_j, \bfc) \right|^2 + \lambda_{\rm bc} \sum_{j=1}^{J_b} | z_j^5 |^2. 
\label{eq:penalized_problem}
\end{equation}
Problems of the form \eqref{eq:penalized_problem} can be solved via Gauss-Newton or Levenberg-Marquardt algorithms; we take the approach of \citet[Appendix C.2]{jalalianDataEfficientKernelMethods2025} and employ Levenberg-Marquardt with Cholesky changes of variables $\bfw_u = L_u\inv \bfz_u$ and $\bfw_g = L_g\inv \bfz_g$, where $K_u(\varphi, \varphi) = L_uL_u\t$ and $K_g(\psi, \psi) = L_gL_g\t$ are the Cholesky factorizations of $K_u(\varphi, \varphi)$ and $K_g(\psi, \psi)$. %

\section{Experimental details}
\label{app:experimental_details}
In the experiment of \cref{sec:numerics}, 
our collocation points $\{(x_j, t_j)\}_{j=1}^{J}$ are the tensor-product of a uniform spatial grid over the interval $[-2s -3,\, s + 3]$ and a uniform time grid over the interval $[0,1]$. We take $N_x = 50$ spatial points and $N_t = 51$ time points, for a total of $J = N_xN_t = 2550$ space-time collocation points. Additionally, the boundary points $\{(x^b_j, t^b_j)\}_{j=1}^{J_b}$ are the tensor product between the same uniform spatial grid and $\{0, 1\}$ for a total of $J_b = 2N_x$ boundary points. 

We take $K_u((x, t), (x', t')) = K_g((x, t), (x', t')) = K_x(x, x')K_t(t, t')$, where $K_x$ and $K_t$ are SPD kernels on $\R$. We choose $K_x$ and $K_t$ to both be Matern kernels, 
\[
K(x, x') = \frac{2^{1 - \nu}}{\Gamma(\nu)} \left( \sqrt{2\nu} \frac{\|x - x' \|}{\sigma} \right) K_\nu \left( \sqrt{2\nu} \frac{\|x - x' \|}{\sigma} \right)  ,
\]
where $\Gamma$ is the Gamma function and $K_\nu$ is the modified Bessel function of the second kind. We take the smoothness $\nu = 5/2$. We set the lengthscale of $K_t$ to be $\sigma_t = 1/\sqrt{N_t}$ and the lengthscale of $K_x$ to be $\sigma_x = 180/Nx$. We initialize the unknowns in \cref{eq:relaxed_problem} at $\bfz_u = \mathbf{0} \in \R^{J(d+1)}$, $\bfz_g = \mathbf{0}\in \R^{J(d+1) + J_b}$, and $\bfc = \mathbf{0} \in \R^{N}$. For the first few iterations of optimization we dynamically adjust the regularization parameters to balance the terms of the loss, ultimately settling on $\lambda_g = 51.8$, $\lambda_{\rm pde} = 2.63 \times 10^{5}$, and $\lambda_{\rm bc} = 6.01 \times 10^4$.  

The ensembles appearing in \cref{fig:learned_interp_1D} and for which the metrics in \cref{tab:sample_metrics}  were computed consist of 1000 particles each and were generated using the forward Euler method with a uniform step-size $\Delta t = 0.01$. 

We make use of the implementation of the GP-PDE approach provided by \citet{jalalianDataEfficientKernelMethods2025}. %
All experiments were run on one Nvidia A100 GPU, although they could be feasibly run on a standard CPU (e.g., on a laptop) as well.

\newpage 
\bibliography{sample.bib}

@article{creswell2018generative,
  title={Generative adversarial networks: An overview},
  author={Creswell, Antonia and White, Tom and Dumoulin, Vincent and Arulkumaran, Kai and Sengupta, Biswa and Bharath, Anil A},
  journal={IEEE signal processing magazine},
  volume={35},
  number={1},
  pages={53--65},
  year={2018},
  publisher={IEEE}
}

@book{berlinet2011reproducing,
  title={Reproducing kernel Hilbert spaces in probability and statistics},
  author={Berlinet, Alain and Thomas-Agnan, Christine},
  year={2011},
  publisher={Springer Science \& Business Media}
}

@article{papamakarios2021normalizing,
  title={Normalizing flows for probabilistic modeling and inference},
  author={Papamakarios, George and Nalisnick, Eric and Rezende, Danilo Jimenez and Mohamed, Shakir and Lakshminarayanan, Balaji},
  journal={Journal of Machine Learning Research},
  volume={22},
  number={57},
  pages={1--64},
  year={2021}
}

@article{cao2024survey,
  title={A survey on generative diffusion models},
  author={Cao, Hanqun and Tan, Cheng and Gao, Zhangyang and Xu, Yilun and Chen, Guangyong and Heng, Pheng-Ann and Li, Stan Z},
  journal={IEEE transactions on knowledge and data engineering},
  volume={36},
  number={7},
  pages={2814--2830},
  year={2024},
  publisher={IEEE}
}

@misc{mateLearningInterpolationsBoltzmann2023a,
  title = {Learning {{Interpolations}} between {{Boltzmann Densities}}},
  author = {M{\'a}t{\'e}, B{\'a}lint and Fleuret, Fran{\c c}ois},
  year = {2023},
  month = may,
  number = {arXiv:2301.07388},
  eprint = {2301.07388},
  primaryclass = {cs, stat},
  publisher = {arXiv},
  doi = {10.48550/arXiv.2301.07388},
  urldate = {2024-10-16},
  archiveprefix = {arXiv}
}

@inproceedings{mauraisSamplingUnitTime2024c,
  title = {Sampling in {{Unit Time}} with {{Kernel Fisher-Rao Flow}}},
  booktitle = {Proceedings of the 41st {{International Conference}} on {{Machine Learning}}},
  author = {Maurais, Aimee and Marzouk, Youssef},
  year = {2024},
  month = jul,
  pages = {35138--35162},
  publisher = {PMLR},
  issn = {2640-3498},
  urldate = {2024-11-26},
  langid = {english}
}

@inproceedings{albergoBuildingNormalizingFlows2022,
  title = {Building {{Normalizing Flows}} with {{Stochastic Interpolants}}},
  booktitle = {The {{Eleventh International Conference}} on {{Learning Representations}}},
  author = {Albergo, Michael Samuel and {Vanden-Eijnden}, Eric},
  year = {2022},
  month = sep,
  urldate = {2025-01-14},
  langid = {english}
}

@misc{albergoStochasticInterpolantsUnifying2023,
  title = {Stochastic {{Interpolants}}: {{A Unifying Framework}} for {{Flows}} and {{Diffusions}}},
  shorttitle = {Stochastic {{Interpolants}}},
  author = {Albergo, Michael S. and Boffi, Nicholas M. and {Vanden-Eijnden}, Eric},
  year = {2023},
  month = mar,
  number = {arXiv:2303.08797},
  eprint = {2303.08797},
  primaryclass = {cond-mat},
  publisher = {arXiv},
  urldate = {2023-03-22},
  archiveprefix = {arXiv},
  langid = {english}
}

@inproceedings{lipmanFlowMatchingGenerative2022,
  title = {Flow {{Matching}} for {{Generative Modeling}}},
  booktitle = {The {{Eleventh International Conference}} on {{Learning Representations}}},
  author = {Lipman, Yaron and Chen, Ricky T. Q. and {Ben-Hamu}, Heli and Nickel, Maximilian and Le, Matthew},
  year = {2022},
  month = sep,
  urldate = {2025-01-14},
  langid = {english}
}

@inproceedings{liuFlowStraightFast2022a,
  title = {Flow {{Straight}} and {{Fast}}: {{Learning}} to {{Generate}} and {{Transfer Data}} with {{Rectified Flow}}},
  shorttitle = {Flow {{Straight}} and {{Fast}}},
  booktitle = {The {{Eleventh International Conference}} on {{Learning Representations}}},
  author = {Liu, Xingchao and Gong, Chengyue and Liu, Qiang},
  year = {2022},
  month = sep,
  urldate = {2025-01-14},
  langid = {english}
}

@misc{chemseddineNeuralSamplingBoltzmann2024,
  title = {Neural {{Sampling}} from {{Boltzmann Densities}}: {{Fisher-Rao Curves}} in the {{Wasserstein Geometry}}},
  shorttitle = {Neural {{Sampling}} from {{Boltzmann Densities}}},
  author = {Chemseddine, Jannis and Wald, Christian and Duong, Richard and Steidl, Gabriele},
  year = {2024},
  month = oct,
  number = {arXiv:2410.03282},
  eprint = {2410.03282},
  primaryclass = {cs, math},
  publisher = {arXiv},
  doi = {10.48550/arXiv.2410.03282},
  urldate = {2024-10-10},
  archiveprefix = {arXiv}
}

@misc{sunDynamicalMeasureTransport2024a,
  title = {Dynamical {{Measure Transport}} and {{Neural PDE Solvers}} for {{Sampling}}},
  author = {Sun, Jingtong and Berner, Julius and Richter, Lorenz and Zeinhofer, Marius and M{\"u}ller, Johannes and Azizzadenesheli, Kamyar and Anandkumar, Anima},
  year = {2024},
  month = jul,
  number = {arXiv:2407.07873},
  eprint = {2407.07873},
  primaryclass = {cs},
  publisher = {arXiv},
  doi = {10.48550/arXiv.2407.07873},
  urldate = {2025-08-12},
  archiveprefix = {arXiv}
}

@misc{wangMeasureTransportKernel2024,
  title = {Measure Transport with Kernel Mean Embeddings},
  author = {Wang, L. and N{\"u}sken, N.},
  year = {2024},
  month = sep,
  number = {arXiv:2401.12967},
  eprint = {2401.12967},
  primaryclass = {math},
  publisher = {arXiv},
  doi = {10.48550/arXiv.2401.12967},
  urldate = {2025-05-01},
  archiveprefix = {arXiv}
}

@misc{beylerConvergenceDeterministicStochastic2025,
  title = {Convergence of {{Deterministic}} and {{Stochastic Diffusion-Model Samplers}}: {{A Simple Analysis}} in {{Wasserstein Distance}}},
  shorttitle = {Convergence of {{Deterministic}} and {{Stochastic Diffusion-Model Samplers}}},
  author = {Beyler, Eliot and Bach, Francis},
  year = {2025},
  month = aug,
  number = {arXiv:2508.03210},
  eprint = {2508.03210},
  primaryclass = {cs},
  publisher = {arXiv},
  doi = {10.48550/arXiv.2508.03210},
  urldate = {2025-08-10},
  archiveprefix = {arXiv}
}

@InProceedings{tsimposOptimalSchedulingDynamic2025,
  title = 	 {Optimal Scheduling of Dynamic Transport},
  author =       {Tsimpos, Panos and Zhi, Ren and Zech, Jakob and Marzouk, Youssef},
  booktitle = 	 {Proceedings of Thirty Eighth Conference on Learning Theory},
  pages = 	 {5441--5505},
  year = 	 {2025},
  editor = 	 {Haghtalab, Nika and Moitra, Ankur},
  volume = 	 {291},
  series = 	 {Proceedings of Machine Learning Research},
  month = 	 {30 Jun--04 Jul},
  publisher =    {PMLR},
  url = 	 {https://proceedings.mlr.press/v291/tsimpos25a.html}
}

@misc{hernandezPDPOParametricDensity2025,
  title = {{{PDPO}}: {{Parametric Density Path Optimization}}},
  shorttitle = {{{PDPO}}},
  author = {Hernandez, Sebastian Gutierrez and Chen, Peng and Zhou, Haomin},
  year = 2025,
  month = may,
  number = {arXiv:2505.18473},
  eprint = {2505.18473},
  primaryclass = {math},
  publisher = {arXiv},
  doi = {10.48550/arXiv.2505.18473},
  urldate = {2025-07-06},
  archiveprefix = {arXiv}
}

@article{chenSolvingLearningNonlinear2021a,
  title = {Solving and Learning Nonlinear {{PDEs}} with {{Gaussian}} Processes},
  author = {Chen, Yifan and Hosseini, Bamdad and Owhadi, Houman and Stuart, Andrew M.},
  year = {2021},
  month = dec,
  journal = {Journal of Computational Physics},
  volume = {447},
  pages = {110668},
  issn = {0021-9991},
  doi = {10.1016/j.jcp.2021.110668},
  urldate = {2025-01-22}
}

@article{kadriOperatorvaluedKernelsLearning2016,
  title = {Operator-Valued {{Kernels}} for {{Learning}} from {{Functional Response Data}}},
  author = {Kadri, Hachem and Duflos, Emmanuel and Preux, Philippe and Canu, St{\'e}phane and Rakotomamonjy, Alain and Audiffren, Julien},
  year = {2016},
  journal = {Journal of Machine Learning Research},
  volume = {17},
  number = {20},
  pages = {1--54},
  issn = {1533-7928},
  urldate = {2024-11-04}
}

@article{alvarezKernelsVectorValuedFunctions2012,
  title = {Kernels for {{Vector-Valued Functions}}: {{A Review}}},
  shorttitle = {Kernels for {{Vector-Valued Functions}}},
  author = {{\'A}lvarez, Mauricio A. and Rosasco, Lorenzo and Lawrence, Neil D.},
  year = {2012},
  month = jun,
  journal = {Foundations and Trends{\textregistered} in Machine Learning},
  volume = {4},
  number = {3},
  pages = {195--266},
  publisher = {Now Publishers, Inc.},
  issn = {1935-8237, 1935-8245},
  doi = {10.1561/2200000036},
  urldate = {2025-08-27},
  langid = {english}
}

@inbook{Owhadi_Scovel_2019, place={Cambridge}, series={Cambridge Monographs on Applied and Computational Mathematics}, title={Optimal Recovery Splines}, booktitle={Operator-Adapted Wavelets, Fast Solvers, and Numerical Homogenization: From a Game Theoretic Approach to Numerical Approximation and Algorithm Design}, publisher={Cambridge University Press}, author={Owhadi, Houman and Scovel, Clint}, year={2019}, pages={154–159}, collection={Cambridge Monographs on Applied and Computational Mathematics}}

@book{ghanem2017handbook,
  title={Handbook of uncertainty quantification},
  author={Ghanem, Roger and Higdon, David and Owhadi, Houman and others},
  volume={6},
  year={2017},
  publisher={Springer New York}
}

@inproceedings{songScoreBasedGenerativeModeling2021,
	title = {Score-{{Based Generative Modeling}} through {{Stochastic Differential Equations}}},
	booktitle = {International {{Conference}} on {{Learning Representations}}},
	author = {Song, Yang and {Sohl-Dickstein}, Jascha and Kingma, Diederik P. and Kumar, Abhishek and Ermon, Stefano and Poole, Ben},
	year = {2020},
	month = oct,
	urldate = {2025-01-14},
	langid = {english}
}

@misc{zhangMeanfieldGamesLaboratory2023,
	title = {A Mean-Field Games Laboratory for Generative Modeling},
	author = {Zhang, Benjamin J. and Katsoulakis, Markos A.},
	year = {2023},
	month = oct,
	number = {arXiv:2304.13534},
	eprint = {2304.13534},
	primaryclass = {cs, stat},
	publisher = {arXiv},
	doi = {10.48550/arXiv.2304.13534},
	urldate = {2024-03-05},
	archiveprefix = {arxiv}
}

@misc{havensAdjointSamplingHighly2025a,
  title = {Adjoint {{Sampling}}: {{Highly Scalable Diffusion Samplers}} via {{Adjoint Matching}}},
  shorttitle = {Adjoint {{Sampling}}},
  author = {Havens, Aaron and Miller, Benjamin Kurt and Yan, Bing and {Domingo-Enrich}, Carles and Sriram, Anuroop and Wood, Brandon and Levine, Daniel and Hu, Bin and Amos, Brandon and Karrer, Brian and Fu, Xiang and Liu, Guan-Horng and Chen, Ricky T. Q.},
  year = 2025,
  month = may,
  number = {arXiv:2504.11713},
  eprint = {2504.11713},
  primaryclass = {cs},
  publisher = {arXiv},
  doi = {10.48550/arXiv.2504.11713},
  urldate = {2025-09-12},
  archiveprefix = {arXiv}
}

@inproceedings{grathwohlFFJORDFreeFormContinuous2018,
	title = {{{FFJORD}}: {{Free-Form Continuous Dynamics}} for {{Scalable Reversible Generative Models}}},
	shorttitle = {{{FFJORD}}},
	booktitle = {International {{Conference}} on {{Learning Representations}}},
	author = {Grathwohl, Will and Chen, Ricky T. Q. and Bettencourt, Jesse and Sutskever, Ilya and Duvenaud, David},
	year = {2018},
	month = sep,
	urldate = {2024-06-14},
	langid = {english}
}

@inproceedings{chenNeuralOrdinaryDifferential2018,
	title = {Neural {{Ordinary Differential Equations}}},
	booktitle = {Advances in {{Neural Information Processing Systems}}},
	author = {Chen, Ricky T. Q. and Rubanova, Yulia and Bettencourt, Jesse and Duvenaud, David K},
	year = {2018},
	volume = {31},
	publisher = {Curran Associates, Inc.},
	urldate = {2024-05-01}
}

@article{mccannConvexityPrincipleInteracting1997,
  title = {A {{Convexity Principle}} for {{Interacting Gases}}},
  author = {McCann, Robert J.},
  year = {1997},
  month = jun,
  journal = {Advances in Mathematics},
  volume = {128},
  number = {1},
  pages = {153--179},
  issn = {0001-8708},
  doi = {10.1006/aima.1997.1634},
  urldate = {2025-08-25}
}

@article{domingo-enrichExplicitExpansionKullbackLeibler2023,
  title = {An {{Explicit Expansion}} of the {{Kullback-Leibler Divergence}} along Its {{Fisher-Rao Gradient Flow}}},
  author = {{Domingo-Enrich}, Carles and Pooladian, Aram-Alexandre},
  year = {2023},
  month = mar,
  journal = {Transactions on Machine Learning Research},
  issn = {2835-8856},
  urldate = {2024-04-09},
  langid = {english}
}

@book{amari2016information,
	title={Information geometry and its applications},
	author={Amari, Shun-ichi},
	volume={194},
	year={2016},
	publisher={Springer}
}

@misc{chenGradientFlowsSampling2023,
  title = {Gradient {{Flows}} for {{Sampling}}: {{Mean-Field Models}}, {{Gaussian Approximations}} and {{Affine Invariance}}},
  shorttitle = {Gradient {{Flows}} for {{Sampling}}},
  author = {Chen, Yifan and Huang, Daniel Zhengyu and Huang, Jiaoyang and Reich, Sebastian and Stuart, Andrew M.},
  year = {2023},
  month = jul,
  number = {arXiv:2302.11024},
  eprint = {2302.11024},
  primaryclass = {cs, math, stat},
  publisher = {arXiv},
  doi = {10.48550/arXiv.2302.11024},
  urldate = {2023-09-28},
  archiveprefix = {arXiv}
}

@article{lasryMeanFieldGames2007,
  title = {Mean Field Games},
  author = {Lasry, Jean-Michel and Lions, Pierre-Louis},
  year = {2007},
  month = mar,
  journal = {Japanese Journal of Mathematics},
  volume = {2},
  number = {1},
  pages = {229--260},
  issn = {1861-3624},
  doi = {10.1007/s11537-007-0657-8},
  urldate = {2024-05-01},
  langid = {english}
}

@misc{jalalianDataEfficientKernelMethods2025,
  title = {Data-{{Efficient Kernel Methods}} for {{Learning Differential Equations}} and {{Their Solution Operators}}: {{Algorithms}} and {{Error Analysis}}},
  shorttitle = {Data-{{Efficient Kernel Methods}} for {{Learning Differential Equations}} and {{Their Solution Operators}}},
  author = {Jalalian, Yasamin and Ramirez, Juan Felipe Osorio and Hsu, Alexander and Hosseini, Bamdad and Owhadi, Houman},
  year = {2025},
  month = apr,
  number = {arXiv:2503.01036},
  eprint = {2503.01036},
  primaryclass = {stat},
  publisher = {arXiv},
  doi = {10.48550/arXiv.2503.01036},
  urldate = {2025-06-20},
  archiveprefix = {arXiv}
}

@misc{domingo-enrichAdjointMatchingFinetuning2025,
  title = {Adjoint {{Matching}}: {{Fine-tuning Flow}} and {{Diffusion Generative Models}} with {{Memoryless Stochastic Optimal Control}}},
  shorttitle = {Adjoint {{Matching}}},
  author = {{Domingo-Enrich}, Carles and Drozdzal, Michal and Karrer, Brian and Chen, Ricky T. Q.},
  year = {2025},
  month = jan,
  number = {arXiv:2409.08861},
  eprint = {2409.08861},
  primaryclass = {cs},
  publisher = {arXiv},
  doi = {10.48550/arXiv.2409.08861},
  urldate = {2025-09-01},
  archiveprefix = {arXiv}
}

@misc{zhangPathIntegralSampler2022,
  title = {Path {{Integral Sampler}}: A Stochastic Control Approach for Sampling},
  shorttitle = {Path {{Integral Sampler}}},
  author = {Zhang, Qinsheng and Chen, Yongxin},
  year = {2022},
  month = mar,
  number = {arXiv:2111.15141},
  eprint = {2111.15141},
  primaryclass = {cs},
  publisher = {arXiv},
  doi = {10.48550/arXiv.2111.15141},
  urldate = {2025-09-01},
  archiveprefix = {arXiv}
}

@article{bernerOptimalControlPerspective2023,
  title = {An Optimal Control Perspective on Diffusion-Based Generative Modeling},
  author = {Berner, Julius and Richter, Lorenz and Ullrich, Karen},
  year = {2023},
  month = oct,
  journal = {Transactions on Machine Learning Research},
  issn = {2835-8856},
  urldate = {2025-09-01},
  langid = {english}
}

@inproceedings{vargasDenoisingDiffusionSamplers2022,
  title = {Denoising {{Diffusion Samplers}}},
  booktitle = {The {{Eleventh International Conference}} on {{Learning Representations}}},
  author = {Vargas, Francisco and Grathwohl, Will Sussman and Doucet, Arnaud},
  year = {2022},
  month = sep,
  urldate = {2024-06-05},
  langid = {english}
}

\newpage
\section*{NeurIPS Paper Checklist}
\begin{enumerate}
\item {\bf Claims}
    \item[] Question: Do the main claims made in the abstract and introduction accurately reflect the paper's contributions and scope?
    \item[] Answer: \answerYes{} %
    \item[] Justification: The abstract succinctly summarizes the contributions of our paper. 

\item {\bf Limitations}
    \item[] Question: Does the paper discuss the limitations of the work performed by the authors?
    \item[] Answer: \answerYes{} %
    \item[] Justification: This is a workshop paper and the results are preliminary. The conclusion discusses avenues for future work. 

\item {\bf Theory assumptions and proofs}
    \item[] Question: For each theoretical result, does the paper provide the full set of assumptions and a complete (and correct) proof?
    \item[] Answer: \answerYes{} %
    \item[] Justification: While we do not have any formal theorems/proofs, the mathematical connections we draw are well-explained and justified. %

    \item {\bf Experimental result reproducibility}
    \item[] Question: Does the paper fully disclose all the information needed to reproduce the main experimental results of the paper to the extent that it affects the main claims and/or conclusions of the paper (regardless of whether the code and data are provided or not)?
    \item[] Answer: \answerYes{} %
    \item[] Justification: Algorithms and hyperparameters are detailed in \cref{sec:numerics,app:optimal_recovery,app:experimental_details}. 

\item {\bf Open access to data and code}
    \item[] Question: Does the paper provide open access to the data and code, with sufficient instructions to faithfully reproduce the main experimental results, as described in supplemental material?
    \item[] Answer: \answerNo{} %
    \item[] Justification: While our code is not ready to be released at this time, we will provide a link to a repository containing code to reproduce the experiments in the camera-ready version. 

\item {\bf Experimental setting/details}
    \item[] Question: Does the paper specify all the training and test details (e.g., data splits, hyperparameters, how they were chosen, type of optimizer, etc.) necessary to understand the results?
    \item[] Answer: \answerYes{} %
    \item[] Justification: See \cref{app:experimental_details}

\item {\bf Experiment statistical significance}
    \item[] Question: Does the paper report error bars suitably and correctly defined or other appropriate information about the statistical significance of the experiments?
    \item[] Answer: \answerNA{} %
    \item[] Justification: Our algorithms and their initializations are deterministic and thus error bars are not needed. For testing the performance of our learned velocities in sampling, we used a large enough ensemble (1000 samples in one dimension) that any statistical errors are very small. 

\item {\bf Experiments compute resources}
    \item[] Question: For each experiment, does the paper provide sufficient information on the computer resources (type of compute workers, memory, time of execution) needed to reproduce the experiments?
    \item[] Answer: \answerYes{} %
    \item[] Justification: See \cref{app:experimental_details}. 
    
\item {\bf Code of ethics}
    \item[] Question: Does the research conducted in the paper conform, in every respect, with the NeurIPS Code of Ethics \url{https://neurips.cc/public/EthicsGuidelines}?
    \item[] Answer: \answerYes{} %
    \item[] Justification: No human subjects were involved in this research. This reseach does not rely on any datasets. 

\item {\bf Broader impacts}
    \item[] Question: Does the paper discuss both potential positive societal impacts and negative societal impacts of the work performed?
    \item[] Answer: \answerNA{} %
    \item[] Justification: The methods proposed are primarily concerned with sampling from probability measures known through their densities, and as such the potential for direct societial impacts is low. 
    
\item {\bf Safeguards}
    \item[] Question: Does the paper describe safeguards that have been put in place for responsible release of data or models that have a high risk for misuse (e.g., pretrained language models, image generators, or scraped datasets)?
    \item[] Answer: \answerNA{} %
    \item[] Justification: 

\item {\bf Licenses for existing assets}
    \item[] Question: Are the creators or original owners of assets (e.g., code, data, models), used in the paper, properly credited and are the license and terms of use explicitly mentioned and properly respected?
    \item[] Answer: \answerYes{} %
    \item[] Justification: We cite the paper corresponding to the code we used for the GP-PDE implementation and will also acknowledge the creators in the camera-ready version. 

\item {\bf New assets}
    \item[] Question: Are new assets introduced in the paper well documented and is the documentation provided alongside the assets?
    \item[] Answer: \answerNA{} %
    \item[] Justification: New code will be released with the camera-ready version. 

\item {\bf Crowdsourcing and research with human subjects}
    \item[] Question: For crowdsourcing experiments and research with human subjects, does the paper include the full text of instructions given to participants and screenshots, if applicable, as well as details about compensation (if any)? 
    \item[] Answer: \answerNA{} %
    \item[] Justification: 

\item {\bf Institutional review board (IRB) approvals or equivalent for research with human subjects}
    \item[] Question: Does the paper describe potential risks incurred by study participants, whether such risks were disclosed to the subjects, and whether Institutional Review Board (IRB) approvals (or an equivalent approval/review based on the requirements of your country or institution) were obtained?
    \item[] Answer: \answerNA{} %
    \item[] Justification: 

\item {\bf Declaration of LLM usage}
    \item[] Question: Does the paper describe the usage of LLMs if it is an important, original, or non-standard component of the core methods in this research? Note that if the LLM is used only for writing, editing, or formatting purposes and does not impact the core methodology, scientific rigorousness, or originality of the research, declaration is not required.
    \item[] Answer: \answerNA{} %
    \item[] Justification: LLMs were not used in any aspect of the methods of this paper.

\end{enumerate}

\end{document}